%% file: paper.tex
\pdfoutput=1
\documentclass[11pt]{article}

\usepackage[]{ACL2023}

\usepackage{times}
\usepackage{latexsym}

\usepackage[T1]{fontenc}
\usepackage[utf8]{inputenc}

\usepackage{microtype}

\usepackage{inconsolata}

\usepackage{amsmath}
\usepackage{amsfonts}
\usepackage{amssymb}

\usepackage{booktabs}
\usepackage{multirow}

\usepackage{paralist}

\usepackage{mdwlist}
\usepackage{graphicx}

\usepackage{color}

\usepackage[ruled,linesnumbered]{algorithm2e}

\usepackage{xspace}

\newcommand{\method}{\textsc{SEScore2}\xspace}
\usepackage{graphicx}
\usepackage{wrapfig}
\usepackage{amsmath,bm,bbm,upgreek}
\usepackage{amsthm}
\usepackage{xcolor}
\usepackage{caption}
\usepackage{subcaption}
\usepackage{floatrow}
\usepackage{url}
\usepackage{booktabs}
\usepackage{multirow}
\usepackage{wrapfig}
\usepackage{floatflt}
\usepackage{lipsum}
\usepackage{makecell}
\usepackage{mathtools}
\usepackage{dblfloatfix}
\usepackage{xspace}
\usepackage{caption}
\usepackage{subcaption}
\usepackage{floatrow}
\usepackage{bbding}
\usepackage{adjustbox}

\newfloatcommand{capbtabbox}{table}[][\FBwidth]

\title{\method: Learning Text Generation Evaluation via Synthesizing Realistic Mistakes}

\author{
  Wenda Xu\textsuperscript{\P},
  Xian Qian\textsuperscript{\dag},
  Mingxuan Wang\textsuperscript{\dag},
  \textbf{Lei Li}\textsuperscript{\P}, 
  \textbf{William Yang Wang}\textsuperscript{\P}
  \\
  \textsuperscript{\dag}ByteDance, \textsuperscript{\P}UC Santa Barbara
  \\
  \texttt{\{wendaxu, leili, william\}@cs.ucsb.edu} 
  \\
  \texttt{\{wangmingxuan.89, qian.xian\}@bytedance.com} 
}

\begin{document}
\maketitle

\begin{abstract}
\input{000abstract}

\end{abstract}

\section{Introduction}
\label{sec:intro}
\input{010intro}

\section{Related Work}
\label{sec:related}
\input{020related}

\label{sec:approach}
\input{025and030}

\section{Experiments}
\label{sec:exps}
\input{040exp}

\section{Conclusion}
\label{sec:conclusion}
\input{050conclusion}

\section*{Limitations}
\label{sec:limit}
\input{060limitation}

\section*{Ethics Consideration}
\label{sec:impact}
\input{070impact}

\bibliography{paper}
\bibliographystyle{acl_natbib}

\newpage

\appendix

\appendix
\label{sec:appendix}
\input{080appendix}
\end{document}

%% file: 000abstract.tex
% Can large-scale raw corpora be utilized to construct general learned metrics? Despite their decent correlation with human judgments, existing learned metrics have limitations, such as model-dependent data synthesis, or are restricted to specific domains or tasks for which human ratings are available. In this paper, we propose \method,  a model-based metric pre-trained on a million-scale synthetic dataset generated through our novel retrieval augmented data synthesis pipeline. Importantly, our unsupervised SEScore2 can outperform COMET and BLEURT, which are supervised on the News domain human ratings, at the TED domain. We evaluate \method over four text generation tasks across three languages and achieve a high degree correlation with human judgments. \method outperforms all prior unsupervised metrics across four text generation evaluation benchmarks, with average Kendall improvements of $0.158$. \method even outperforms SOTA supervised BLEURT at WebNLG20 and BAGEL benchmark

Is it possible to train a general metric for evaluating text generation quality without human-annotated ratings? Existing learned metrics either perform unsatisfactorily across text generation tasks or require human ratings for training on specific tasks. In this paper, we propose \method, a self-supervised approach for training a model-based metric for text generation evaluation. The key concept is to synthesize realistic model mistakes by perturbing sentences retrieved from a corpus. The primary advantage of the \method is its ease of extension to many other languages while providing reliable severity estimation. We evaluate \method and previous methods on four text generation tasks across three languages. \method outperforms unsupervised metric PRISM on four text generation evaluation benchmarks, with a Kendall improvement of $0.078$. Surprisingly, \method even outperforms the supervised BLEURT and COMET on multiple text generation tasks. The code and data are available at \url{https://github.com/xu1998hz/SEScore2}\footnote{Part of the work is done while WX is an intern at ByteDance.}.

% Can we train a general metric to evaluate the quality of text generation without human annotated ratings? Existing learned metrics are either unsatisfactory across text generation tasks, or require human ratings for training on specific tasks. 
% In this paper, we propose \method, a self-supervised method to train a model-based metric for text generation evaluation. 
% The key idea is synthesizing realistic model mistakes based on perturbating raw sentences retrieved from a corpus.
% We evaluate \method and previous methods over four text generation tasks across three languages. \method outperforms all prior unsupervised metrics across four text generation evaluation benchmarks, with an average Kendall improvement of $0.158$. Surprisingly, \method even outperforms the supervised BLEURT and COMET on multiple text generation tasks\footnote{All codes and data will be released upon camera ready.}.

%% file: 010intro.tex
Recently, researchers made significant progress in text generation: translation \cite{625af6f84b724fcda0a8bf6026cc922f}, structured data-to-text \cite{gardent-etal-2017-webnlg}, dialogue generation \cite{https://doi.org/10.48550/arxiv.1506.05869}, and summarization \cite{chopra-etal-2016-abstractive}. Automatic metrics are essential for the development of text generation models as they replace expensive human labor and are able to evaluate the generation performance \cite{https://doi.org/10.48550/arxiv.2006.14799}, as well as guide the generation process \cite{JauregiUnanue2021BERTTuneFN, freitag-etal-2022-high}. How can we efficiently and effectively train a metric for general text generation tasks?

Depending on the inputs, we can categorize evaluation metrics into source-based, hybrid-based, and reference-based metrics. 
Source-based metrics estimate text quality through the source and are useful when reference is noisy or unavailable \cite{louis-nenkova-2013-automatically, kepler-etal-2019-openkiwi}, but they may produce sub-optimal results and explore spurious correlations \cite{durmus-etal-2022-spurious}. Reference-based metrics, when paired with high-quality references, can reflect text generation quality, regardless of source modalities (e.g audio and triples). Hybrid metric COMET \cite{rei-etal-2020-comet} uses both source and reference. In this work, we aim to construct a reference-based metric, as it is invariant to the source modality, making it suitable for use across various tasks.

Although learned metrics have been shown to be more effective than rule-based metrics (e.g BLEU \cite{papineni-etal-2002-bleu}), they still have limitations in terms of evaluation capability and applicability to specific tasks. Supervised metrics such as BLEURT \cite{sellam-etal-2020-bleurt} and COMET \cite{rei-etal-2020-comet} are superior in evaluation, but restricted to tasks with human ratings of generated text. Unsupervised metrics, such as BERTScore \cite{https://doi.org/10.48550/arxiv.1904.09675} and BARTScore \cite{https://doi.org/10.48550/arxiv.2106.11520}, do not require human ratings for training, but their correlation with human judgment on specific tasks is still inferior compared to the best supervised metrics \cite{freitag-etal-2021-results}.
% Recent research has incorporated learned components into evaluation, including \textbf{supervised learned metrics} that are directly optimized from human ratings \cite{rei-etal-2020-comet, sellam-etal-2020-bleurt} and \textbf{unsupervised learned metrics}, which obtain training objectives like MLM pretraining \cite{} or sequence-to-sequence pretraining \cite{thompson-post-2020-automatic, https://doi.org/10.48550/arxiv.2106.11520}.

\begin{figure}
    \centering
    \includegraphics[width=\linewidth]{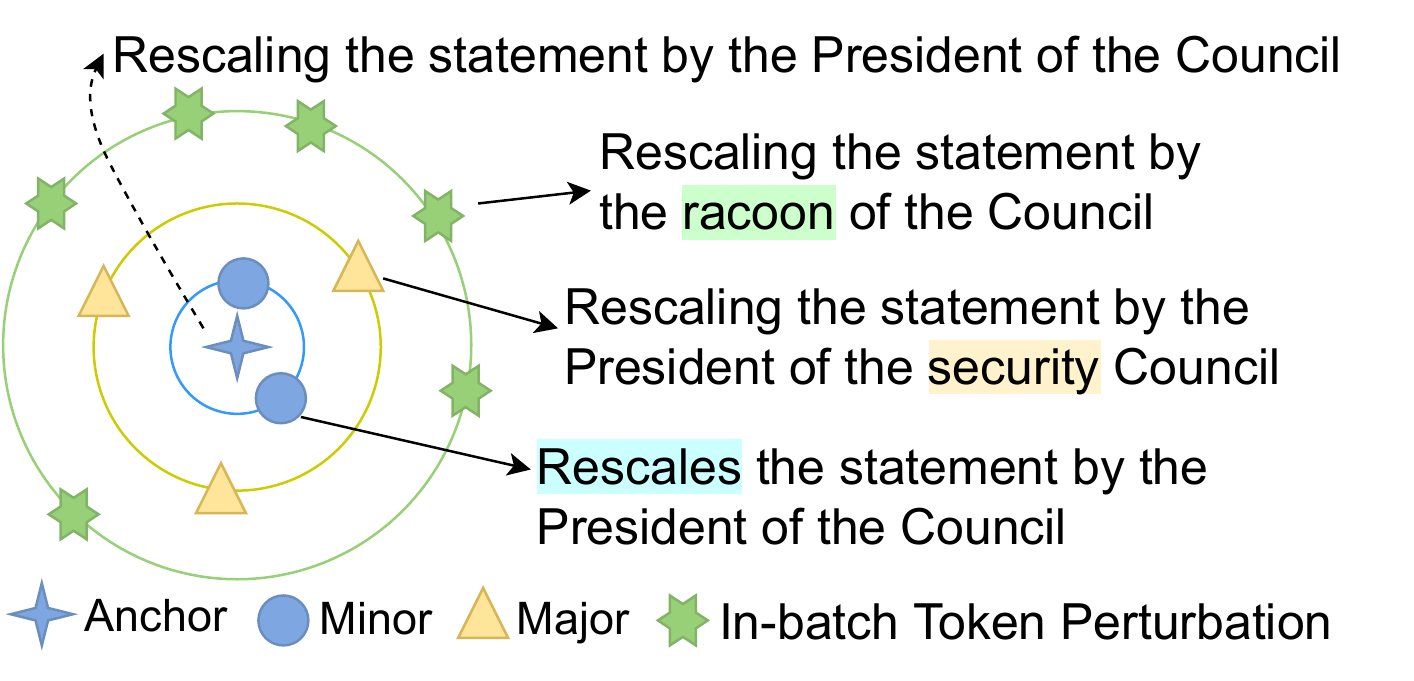}
    \caption{4-point star represents the anchor sentence. Circles and triangles represent the sentences with minor and major mistakes. Both are hard negatives. Green stars are easy negatives produced by random token transformations. Circles that are inner indicate the negative samples that are harder.}
    \label{fig:teaser}
\end{figure}

Our goal of this paper is to devise a reference-based automatic evaluation metric that 1) can be learned without a human quality score, 2) align well with human judgments, and 3) can be generalized to a wide variety of domains and NLG tasks. To achieve this, we propose a self-supervised training method using text with synthetic mistakes. Our main intuition is that these synthetic mistakes should contain errors at different severity levels and appear realistic ("realistic mistakes” are defined as natural and model output-like mistakes). We are inspired by the human evaluation protocol MQM \cite{Freitag_2021} which assesses translation quality by identifying errors with two levels of severity. To make the synthetic mistakes realistic, we mine surface differences among groups of similar sentences and use these different phrases to perturb the original text and construct mistakes (See Figure \ref{fig:template}). Unlike previous methods that utilize generative models to synthesize errors \cite{xu-etal-2022-not}, our approach employs retrieved similar sentences, making it more general. To encompass text diversity, anchor texts are sampled from large-scale parallel corpora. Additionally, a novel pretraining signal is proposed to train \method, which aims to resemble the way humans grade model outputs by estimating the severity levels of each mistake \cite{Freitag_2021}. To date\footnote{May 25, 2023}, we support six languages: English, German, Chinese, Japanese, Russian, Spanish. The primary advantage of the \method is its ease of extension to numerous other languages while providing reliable severity estimation. Our contributions to this paper are as follows:
\begin{itemize}
    \item We propose \method, a self-supervised (SSL) method to train a metric for general text generation tasks without human ratings; 
    \item We develop a technique to synthesize candidate sentences with varying levels of mistakes for training. To make these self-constructed samples realistic, we introduce retrieval augmented synthesis on anchor text;
    \item We annotate an additional human rating dataset for WMT21 German-to-English testing set fol- lowing MQM human annotation procedure and we release it for public use;
    \item Our experiments demonstrate that \method is effective in a wide range of NLG tasks and surpasses the top unsupervised metrics PRISM by $0.078$. Additionally, it also outperforms or matches the supervised metrics in terms of Kendall correlation.
\end{itemize}

% novel retrieval-augmented data synthesis pipeline, which models surface form differences between text pairs to simulate compositional errors in the model outputs. This data construction process is independent of any generative models to prevent biases toward model-specific errors. 

%% file: 020related.tex
% The traditional \textbf{rule-based metrics} are commonly relied on hand-crafted features. They define a set of heuristic rules to estimate the model quality by either counting the number of n-gram matchings (BLEU \cite{papineni-etal-2002-bleu} chrF \cite{popovic-2015-chrf}, METEOR \cite{banerjee-lavie-2005-meteor}, ROUGE \cite{lin-2004-rouge}) or computing edit distances between candidates and human references (e.g TER \cite{snover-etal-2006-study}). However, the rule-based metrics cannot go beyond lexical differences between reference and candidates. 
Human evaluation metrics such as crowd-worker evaluation using direct assessment (DA) are widely used in WMT shared task competition \cite{ma-etal-2018-results, ma-etal-2019-results, mathur-etal-2020-results}.  \citet{mathur-etal-2020-results, Freitag_2021} find that crowd-workers fail to discriminate human and machine outputs. \citet{Freitag_2021} improves human ratings by using Multidimensional Quality Metrics (MQM) framework \cite{lommel2014mqm} with language experts. Each annotated error can be categorized into multiple types and is associated with different severity levels, such as major and minor. 

Automatic evaluation metrics such as rule-based metrics (e.g. n-gram matching BLEU \cite{papineni-etal-2002-bleu}, chrF \cite{popovic-2015-chrf}) and distance-based (e.g. TER \cite{snover-etal-2006-study}) have been commonly used in text generation evaluations because they are fast and domain invariant. However, they have limitations in capturing semantics and long-distance dependencies \cite{https://doi.org/10.48550/arxiv.1904.09675}. The supervised learned metrics \cite{rei-etal-2020-comet, sellam-etal-2020-bleurt} are directly optimized from human ratings. However, they may have poor generalization to unseen domains and tasks \cite{freitag-etal-2021-results}. Unsupervised learned metrics attempt to obtain training objectives other than human ratings \cite{https://doi.org/10.48550/arxiv.1904.09675, zhao-etal-2019-moverscore, thompson-post-2020-automatic, https://doi.org/10.48550/arxiv.2106.11520}. However, as pointed out by \cite{Freitag_2021}, they are limited on the error types that they can evaluate (e.g accuracy) and can not go beyond (e.g fluency or style). Some recent studies attempt to mitigate this issue by generating synthetic data via paraphrasing and perturbations \cite{sellam-etal-2020-bleurt, kryscinski-etal-2020-evaluating, gao-etal-2021-simcse}. To further derive the fine-grained pretraining signals, SEScore \cite{xu-etal-2022-not} leverages language models to generate multiple error types in one segment and estimate each error's severity level. However, model-dependent data synthesis can intrinsically introduce model bias and limit the diversity of data samples. 

\method develops a novel retrieval augmented synthesis technique, which is task-agnostic and can simulate diverse and realistic model mistakes using the parallel corpora. Inspired by \citet{Freitag_2021}, we obtain our pretraining signals aligning with human grading process.

%We strengthen this argument by demonstrating our unsupervised \method can outperform supervised metrics, which are trained on News human ratings in TED domain.   
% \noindent\textbf{Source-based Metrics}

% \noindent\textbf{Reference-based Metrics}

% \noindent\textbf{Human Evaluation}

%% file: 025and030.tex
\begin{figure*}
    \centering
    \includegraphics[width=.98\linewidth]{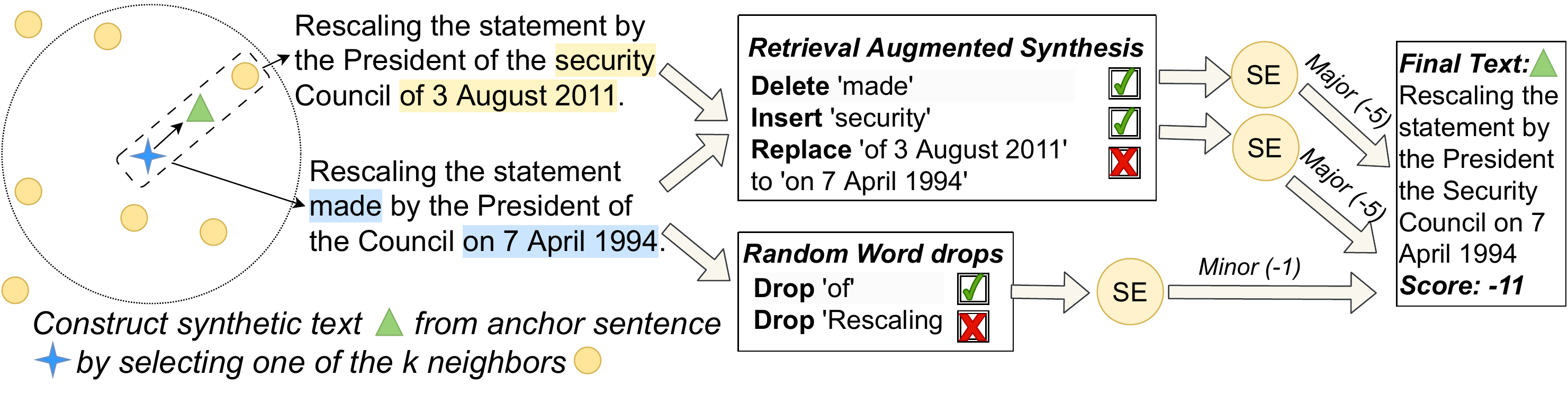}
    \caption{Retrieval Augmented Synthesis: we denote anchor text, selected neighbor, and synthesized text as blue star, circle and triangle respectively. We randomly select a subset of proposed transformations (ticks) and estimate severity measures (SE) on them. Final score sums up  the individual severity measures. }
    \label{fig:template}
\end{figure*}

\section{Problem Definition}
The task is to estimate the quality score between a reference and a model-generated hypothesis. Follow
\citet{Freitag_2021}, if the model error semantically alters the meaning of the sentence, we label it as major, otherwise as minor. See the example in Figure \ref{fig:teaser}, where a 4-point star represents the reference, and circles and triangles represent the sentences with minor and major mistakes, respectively.
A minor mistake contributes a $-1$ score, and a major mistake contributes a $-5$ score. The severity score of a hypothesis is the sum of all minor and major contributions and is no less than $-25$.

Given a set of reference, hypothesis, human-annotated severity score triples ($\mathbf  x$, $\mathbf y$, $\mathbf s$), our goal is to train a learned metric, $M(\mathbf x, \mathbf y) \rightarrow s$. Due to the scarcity of human ratings, such triples are unavailable for most tasks. 
Thus, we propose an automatic way to generate synthetic training samples.

% Our approach has two components: retrieval-based hypothesis generation ($y$) for each anchor sentence ($x$) and automatic severity score labeling ($s$).

\section{The \method Approach}
\method is a SSL technique, initialized with pretrained embeddings, like BERT, then trained with task-agnostic NLG evaluation objective on large-scale synthetic data. No specific fine-tuning is required at inference time evaluation. There are three steps: 1) data construction that samples source-target pairs $(\mathbf{t}, \mathbf{x})$ from  machine translation (MT) corpora and creates synthetic text $\mathbf{y}$ from $\mathbf{x}$ using retrieval augmented synthesis; 2) severity score labeling that automatically generates label $\mathbf{s}$ using $(\mathbf{t}, \mathbf{y})$; 3) model training which pretrains a regression model $M$ using triples $(\mathbf{x}, \mathbf{y},\mathbf{s})$. 
%Parallel MT corpora are only required during data construction. 
During inference, \method only takes reference and model output as input and estimates the quality score, which can be applied to different text generation tasks. 
Detailed implementations of the model can be found in Appendix \ref{sec:quality_model}. 

\subsection{Retrieval Augmented Synthesis} 
Given a reference, a common way to generate a negative hypothesis is in-batch negative samplings or in-batch token insertions or replacements \cite{fu-etal-2022-contextual}. However, as shown in Figure \ref{fig:teaser}, these approaches mainly produce negative samples that are syntactically or semantically incorrect, which are not the case for modern text generation models \cite{Freitag_2021}, see Figure \ref{fig:teaser}.  Therefore training with these samples could not help distinguish model-generated hypotheses. 

Thus we proposed to use retrieval-based approaches to search negative samples. More specifically, given a text corpus, \method finds the $k$ nearest neighbors of $\mathbf{x}$ based on their vector representation using pretrained language embeddings (ex. LASER\footnote{We used LASER3, which supports over 200 languages.}).

We control the text pair proximity by setting a margin criterion when retrieving the $k$ nearest neighbors \cite{schwenk-etal-2021-wikimatrix}. For fast k-NN search, we use an index table. Details refer to Appendix \ref{sec:index_table}. Based on our preliminary study, the margin criterion ($m=1.06$) can retrieve sentences with similar text structure or semantics. Detailed numbers can be found in the Appendix \ref{sec:laser_setting}.  

%To guarantee the syntactical and semantical quality of the negative samples,  
We did not always use the first nearest neighbor because, in many cases, they are too close to the anchor. To increase a diverse coverage of errors, we randomly pick one of those $k$ nearest neighbors $\mathbf z$ (All circles within the loop have equal chances to be selected in Figure \ref{fig:template}). We use edit distance algorithm \cite{snover-etal-2006-study} to decompose the surface form difference between $\mathbf x$ and $\mathbf z$ into a chain of perturbations such as insertion, deletion, and replacement $\mathbf z = P_n(P_{n-1}(\dots(P_1(\mathbf x))))$. In addition, we include random word drops to diversify the errors further. Each $P_i$ is a candidate perturbation to be applied to the text $\mathbf x$. 

According to the human evaluation study \cite{Freitag_2021}, models are most likely to produce fewer than 5 errors. Otherwise, they are labeled as catastrophic errors.
%For models that can produce more than five errors, they will be labeled catastrophic errors with the lowest human ratings. To align this annotation procedure with our error synthesis process,  we first randomly select 5 perturbations from $n$ for each sentence $x$. Next, we randomly select $1$ to $5$ perturbations from the pre-selected five candidates such that each sentence can contain compositional errors. 
Thus for each $\mathbf x$, we randomly select 5 out of the $n$ perturbations. Each time, we apply a random subset of the five perturbations to transform $\mathbf x$ to a negative sample $\mathbf y$, which contains no more than 5 compositional errors.
%Lastly, we can use these perturbations to create negative samples in a linear fashion to transform $x$ into a negative sample $y$ and $y$ is an interpolation between $x$ and $z$. 
One challenge is synthesizing positive samples for the anchor since no ground truth is available. Inspired by \cite{gao-etal-2021-simcse}, we leverage the dropout function to simulate paraphrasing embeddings by feeding the anchor twice. In addition, In-batch negatives are used to approximate the catastrophic errors. 
%Lastly, in-batch negatives are included for the similarity lower bound (score: $-50$)

%In Figure \ref{fig:template}, we demonstrate the data synthesis process from sentence pair $(x, z)$ and show an example of synthetic text $y$ that is constructed from the proposal. 
In figure \ref{fig:template}, we demonstrate that our retrieval augmented synthesis can synthesize errors that are contextually sound but semantically or syntactically deviate from the reference. For example, drop of "of" introduces syntactical error whereas modifications of "made" and "security" introduces semantic errors.

% \ww{Where did you discuss calibration or rounding?}

\subsection{Automatic Severity Score Labeling}
Once we get the synthesized text $\mathbf{y}$, we need to label its severity score. We design severity functions for all types of perturbations, and \textbf{the score of a synthesized text $\mathbf{y}$ is the sum of all severity estimations}.
Inspired by the human evaluation \cite{Freitag_2021}, we consider two levels of severity measures: major (score: $-5$) and minor (score: $-1$), for each error in the candidate outputs. An error is major if it alters the core meaning of the sentence. See triangle and circle examples in Figure \ref{fig:teaser}. Each perturbation is estimated independently to avoid the influence of the others. $\mathbf{t}$ is the machine translation pair of $\mathbf{x}$. $\mathbf{t}$ and $\mathbf{x}$ will be used for insertion/replacement severity estimation.

For insertion and replacement, the severity score is determined by the likelihood of the inserted or replaced tokens. We use a cross-lingual MLM model such as XLM \cite{NEURIPS2019_c04c19c2} to estimate the likelihood. The intuition is that XLM with TLM can model co-occurrences and alignments between source-target tokens. If an error alters the meaning, XLM will be unlikely to restore altered tokens in the perturbed location under the MT source sentence $\mathbf{t}$ and the rest of $\mathbf{y}$'s contexts. 

The severity estimation of a single perturbation $P_i$ on $x$ to $\mathbf y_i$ can be decomposed into two steps: In the first step, we replace perturbed tokens of $\mathbf y$ with masked tokens. Let $\mathbf y_{\mathrm{mask}}$ denote the masked text after the first step, and  $\mathbf m$ denotes the perturbed tokens with length $l$, probability $p_{\mathrm{insert,replace}}=\frac{1}{l}\sum_{i=1}^{l} P(\mathbf m_i|\mathbf t, \mathbf y_{\mathrm{mask}})$ represents the likelihood of this insertion or replacement. For the delete operation, we use TF-IDF weight $w_{\mathrm{delete}}$ to approximate the salience of deleted tokens. The importance weights of the deleted tokens are formulated as, $w_{delete}=max(w_i)$, $i = {1, ..., l}$ with $l$ tokens. Our intuition has two parts: First, we assume that delete operation creates a minor error if all deleted words lack meaningful semantics (e.g. stop words). Second, if one word in the word span has high importance weights (TF-IDF), deletion of the whole word span will alter the sentence semantics. Therefore, we find the maximum of $w_i$ for $w_{delete}$, instead of mean.  

Lastly, an operation is minor if $p_{\mathrm{insert,replace}} \ge \gamma$ or $w_{\mathrm{delete}} < \lambda$, and major otherwise.
where $\gamma$ and $\lambda$ are predefined threshold\footnote{$\lambda=1$ and $\gamma=0.1$ for all three languages. Discussions of hyperparameter choices are included in Appendix \ref{sec:choices_threshold}}. 
% ($P_i(\mathbf x) \rightarrow \mathbf y_i$)

%we adopt a cross-lingual MLM model to perform the severity estimation, such as XLM \cite{NEURIPS2019_c04c19c2}. 

%We first mask the perturbed locations of text $\mathbf{y}$ to obtain $\mathbf{y}_{mask, i}$ where $y_i$ is produced by a single text transformation $P_i$ of $x$. 

%To formulate this estimation,  our severity estimation model will take in three inputs: MT source tokens $t=(t_0, ..., t_m)$; synthesized text tokens with mask, $y_{mask,i}=(y_0,...,m,...y_n)$, where perturbed locations are masked; masked tokens $m=(m_0, ..., ,m_j)$. Then, XLM takes in concatenated source text $t$ with $y_{mask, i}$ and estimate the probability $p$ to recover span $m$ from masks (Figure \ref{fig:severity_measure} shows a single text transformation and its corresponding severity estimation procedure). 

%By setting a threshold $\gamma$, we assign minor label to the transformation $T_i$ if $p_{mask}>=\gamma$ and severe if $p_{mask}<\gamma$. 

% In this case, 
%we use TF-IDF weights of each token to approximate their salience. 

%By setting a threshold $\lambda$, if the importance weights $w$ of the modified span, $w<=\lambda$, the transformation $T_i$ is minor and if $w>\lambda$, error is severe\footnote{$\lambda=1$ and $\gamma=0.1$ for all three languages.}.

% The severity measure is estimated independently on each perturbation and the pretraining signal for each sentence is the sum of all severity measures. 

\begin{figure}
    \centering
    \includegraphics[width=\linewidth]{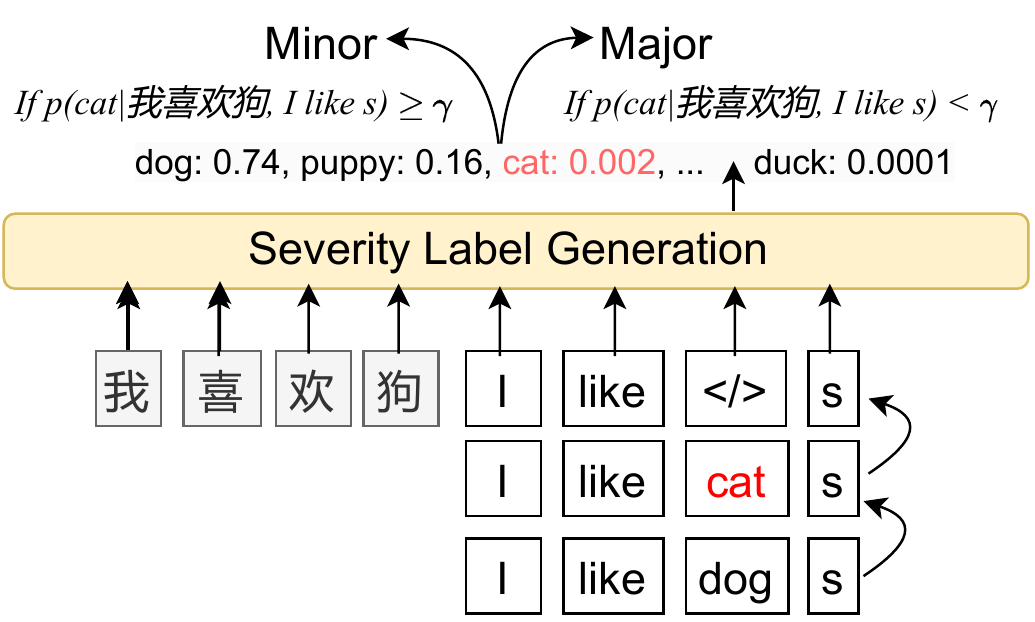}
    \caption{Source Chinese text means 'I like dogs'. First, our retrieval augmented synthesis replaces 'dog' with 'cat'. Then, 'cat' is replaced by a special token '</s>' and we estimate the probability of recovering '</s>'  to 'cat' given the source and target context. Then, we apply a threshold to generate major and minor labels.}
    \label{fig:severity_measure}
\end{figure}

% \subsection{Calibration on the Synthetic Data}
% Besides to precisely rank two sentences (better and worse), \method needs to have the ability to distinguish sentence pairs that are tied - Human ratings of two sentences are the same. One intuitive approach is to round up the precision of the regression output so that two sentences receive the same score. However, without the access of the ground truth distributions, blindly control the precision of the regression can hurt its ranking ability.  

\section{Quality Prediction Model}
\label{sec:quality_model}
We initialized our quality prediction model with pretrained masked language model (e.g. XLM-R). During training, \method takes in text $x$ and synthesized text $y$, supervised with regression score $s$. We drive the sentence embedding from the average pooling of the last layer. Inspired by the prior approach \cite{shimanaka-etal-2018-ruse}, we extract two features between sentence embeddings $x$ and $y$: 1) element-wise sentence product and 2) element-wise sentence difference. We concatenated above two features into one vector and feed into a neural network regressor and trained with mean squared error. During inference, when given unseen candidate and reference pair, \method can directly output a regression score.

%% file: 040exp.tex
% \ww{Maybe I'm missing this. But it seems that your method might be sensitive to the choice of $\lambda$. I wonder if you can elaborate more about how to choose $\lambda$ in the method section, and if you can show ablations on $\lambda$ to tell us how sensitive it is.}

To verify the generalization ability of the \method, 
we investigate the following questions: 
\begin{itemize}
\item Can \method be generalized to multiple domains of the same task? 
\item Can \method's language checkpoint X be used to evaluate all languages Y to X's outputs?
\item Can \method's language checkpoint X be used to evaluate all different text generation tasks on language X? 
\item How to interpret \method?
\end{itemize}
Corresponding to the aforementioned evaluation aspects:
\begin{itemize}
\item We test \method over two different domains (News and TED) at WMT. 
\item We test \method over multiple Y-to-English directions. 
\item We test \method's English checkpoint over a diverse set of NLG tasks: Machine Translation, Speech Translation, Data-to-Text, and Dialogue Generation. 
\item We test \method over multiple evaluation dimensions. Moreover, we conduct comprehensive experiments for each component of \method and analyze the leading factors contributing to the final result. 
\end{itemize}

\subsection{Pretraining Step}

\subsubsection{Pretraining Data}
\label{sec:pretraining_data_stats}
We collected our pretraining data from WMT17-19 publicly available datasets. Details of data collections can be found in the Appendix \ref{sec:pretrain_data_construct}. We randomly sampled 5M, 4.5M, and 5M sentence pairs for Zh-En, En-De, and En-Ja respectively. We use each target language sentence as an anchor to retrieve the 128 nearest neighbors to build the index table and use parallel sentences to compute severity measures. We train separate checkpoints for each language direction and we use the final English checkpoint to evaluate \method in different text generation tasks. To ensure diversity, our index table includes collected WMT News domain data and Wikipedia dumps (See Table \ref{tab:data_stats} for details). We use WMT20 Zh$\rightarrow$En and En$\rightarrow$De with MQM labels \cite{Freitag_2021} as our development sets.

\begin{table}
\resizebox{\textwidth}{!}{% 
\capbtabbox{
        \resizebox{\textwidth}{!}{%  
        \begin{tabular}{@{}lllll@{}}
            \toprule \multicolumn{1}{c}{} & \multicolumn{2}{c}{\bf Index Table} & \multicolumn{2}{c}{\bf Pretraining Data} \\
            \cmidrule(r){2-3}
            \cmidrule(r){4-5}
            \multicolumn{1}{c}{Language} & News & Wikipedia & Anchor & Retrieved \\ \midrule
            \multicolumn{1}{c}{English} & 20M & 20M & 5M & 13.5M\\
            \multicolumn{1}{c}{German} & 4.5M & 16M & 4.5M & 13.2M\\
            \multicolumn{1}{c}{Japanese} & 18M & 12M & 5M & 13.3M\\
            \bottomrule
        \end{tabular}
        }
}{
 \caption{Statistics for Index table and pretraining data. }
 \label{tab:data_stats}
}
}
\end{table}

\begin{table*}[t]\small
    \centering
    \resizebox{\textwidth}{!}{%  
    \begin{tabular}{@{}l|cccccccc@{}}
        \toprule
        \multicolumn{2}{c}{} & {\bf MT(Zh$\rightarrow$En)} & {\bf MT(De$\rightarrow$En)} & {\bf MT(En$\rightarrow$De)} & {\bf ST(En$\rightarrow$Ja)} & {\bf D2T(En)} & {\bf Dialog(En)} & {\bf Overall}\\ \midrule
        \parbox[t]{5mm}{\multirow{2}{*}{\rotatebox[origin=c]{90}{\shortstack{With}}}}
        & \multicolumn{1}{c}{BLEURT} & 0.291 & 0.266 & 0.252 & 0.463 & 0.168 & 0.229 & 0.278 \\
        & \multicolumn{1}{c}{COMET(DA)} & 0.290  & 0.250 & 0.249 & 0.405 & - & - & -\\
        \cmidrule(l){1-9}
        \parbox[t]{5mm}{\multirow{7}{*}{\rotatebox[origin=c]{90}{\shortstack{Without Supervision}}}}
        
        & \multicolumn{1}{c}{TER} & 0.173 & -0.046 & 0.115  & -0.082 & -0.090 & -0.087 & -0.003\\
        & \multicolumn{1}{c}{BLEU} & 0.134 & 0.068 & 0.098 & 0.202 & 0.084 & 0.109 & 0.116\\
        & \multicolumn{1}{c}{ChrF} & 0.158 & 0.074 & 0.130 & 0.240  & 0.094 & 0.108 & 0.134\\
        & \multicolumn{1}{c}{BARTScore} & 0.208 & 0.047 & 0.042 & -0.123 & 0.113 & 0.203 & 0.082\\
        & \multicolumn{1}{c}{BERTScore} & 0.248 & 0.205 & 0.179 & 0.213 & 0.154 & 0.171 & 0.195\\
        & \multicolumn{1}{c}{PRISM} & 0.240 & 0.174 & 0.215 & 0.198 & 0.163 & 0.217 & 0.201\\
        & \multicolumn{1}{c}{SEScore} & 0.281 & 0.249 & 0.226 & 0.361 & 0.155 & 0.205 & 0.246\\
        & \multicolumn{1}{c}{\method} & \textbf{0.310} & \textbf{0.250} & \textbf{0.243} & \textbf{0.458} & \textbf{0.182} & \textbf{0.233} & \textbf{0.279}\\
        \bottomrule
    \end{tabular}
    }
    \caption{Segment-level Kendall correlation on En-De, De-En and Zh-En for WMT21, En-Ja for IWSLT22, WebNLG20 data-to-text and BAGEL dialogue generation. \method significantly outperforms all unsupervised metrics in all tasks and BLEURT in D2T and dialogue generation based on William's pair-wise significance test, with p values < 0.05. We bold the best performed unsupervised metrics. }%\ww{Supervised vs. Unsupervised?}}
    \label{tab:main-text-generation}
\end{table*}

\subsubsection{Scoring Model}
 We use Rembert \cite{chung2020rethinking} as the backbone for all text generation tasks (other backbone choices are discussed in the Appendix \ref{sec:model_init}). We use Adam optimizer and set batch size, learning rate, and dropout rate to 256, 3e-5, and 0.1 respectively. We use the mean squared error to train the metric model. All checkpoints from Rembert trained for 15,000 iterations. We use 8 A100 GPUs to train for 18 hours for each checkpoint. 

\subsection{Baseline Model}
For all NLG tasks, we include 1) three n-gram, distance-based baseline metrics: BLEU \cite{papineni-etal-2002-bleu}, ChrF \cite{popovic-2015-chrf} and TER \cite{snover-etal-2006-study}; 2) four best performed learned metrics without human ratings: PRISM \cite{thompson-post-2020-automatic}, BARTScore \cite{https://doi.org/10.48550/arxiv.2106.11520}, BERTScore \cite{https://doi.org/10.48550/arxiv.1904.09675} and SEScore \cite{xu-etal-2022-not}; and 3) two SOTA supervised metrics: COMET \footnote{Since COMET is a source-reference-based approach only applicable to translation tasks, we only used to generate results for machine and speech translation} and BLEURT. Implementation details are discussed in Appendix \ref{sec:baseline_implmentations}.

\subsection{Evaluation Procedure}
For all the text generation tasks, we compute segment-level Kendall correlation between metric outputs and human scores (We include all Spearman correlation in the Appendix Table \ref{tab:dialogue-multi-spearman}, \ref{tab:full-lang-results-spearman} and \ref{tab:main-text-generation-spearman}. They yield the same conclusion as Kendall correlation). We conduct William's pair-wise significance test \cite{graham-baldwin-2014-testing} to highlight the significant improvements. 

\paragraph{Machine Translation} 
For En-De and Zh-En, we used publicly available WMT21 News and TED's human annotations \cite{freitag-etal-2021-results}. We also hired 3 professional linguists to annotate 1000 testing samples from WMT21 De-En News domain using MQM human evaluation procedures \cite{Freitag_2021}. Detailed testing statistics are in Appendix Table \ref{tab:test_data_stats} and detailed human annotation procedures are in Appendix \ref{sec:human_annotations}.  

\paragraph{Dialogue Generation} Public BAGEL benchmark contains target utterance generation for spoken dialogue systems. This benchmark contains 202 model outputs. Each sample is annotated in the aspect of naturalness, informativeness, and quality. 

\paragraph{Data-to-Text Generation} Public WebNLG2020 \cite{zhou-lampouras-2020-webnlg} contains 17 models and each contains 177 outputs. Each sample is annotated by five aspects: correctness, data coverage, fluency, relevance, and text structure.

\paragraph{Speech-to-Text} We use IWSLT22 English-to-Japanese (En-Ja) human annotations. The benchmark contains four systems and each contains 118 outputs. All human annotations were done using JTF MQM variant \cite{JTF}.

\subsection{Overall Performance}
In Table \ref{tab:main-text-generation}, we demonstrate metrics' overall performance in machine translation, speech translation, data-to-text, and dialogue generation. \method outperforms the best rule-based metric chrF (Kendall=$0.134$) significantly in the overall Kendall correlation, with Kendall improvement of $0.145$. \method outperforms all unsupervised learned metrics significantly in all four text generation tasks and three MT translation directions. In particular, \method outperforms PRISM (Kendall=$0.201$) with Kendall improvement $0.078$. More importantly, \method outperforms the supervised BLEURT in two of the four text generation tasks and achieves a higher Kendall correlation overall across four tasks, with Kendall improvement of $0.014$ in D2T(En) and $0.004$ in Dialog(En). 

\subsection{\method achieves consistent superior performance for different text generation tasks}
\label{sec:task-study}
For Machine Translation, \method outperforms all unsupervised metrics significantly across all three language directions. Despite all language directions being present in the training sets of both BLEURT and COMET, \method outperforms both supervised metrics COMET and BLEURT in Zh-En and achieves comparable performance to COMET and close performance to BLEURT at En-De and De-En. For speech translation, \method outperforms all unsupervised metrics significantly and leads COMET by a large margin. One explanation for this improvement could be that human ratings for English to Japanese were not included in COMET's training data, highlighting the limitations of supervised metrics in unknown domains or language directions. Lastly, \method outperforms all supervised and unsupervised metrics at data-to-text and dialogue generation. Compared to BLEURT, which is supervised by translation human rating, \method can achieve superior generalization capability in non-translation tasks, such as data-to-text and dialogue generation.

\subsection{SEScore2 achieves consistent superior performance for translation into the same target languages}
SEScore2 is consistently better on a variety of text generation tasks with the same generation language. For the machine translation task, we further investigate \method's generalization capabilities over different languages to X translation outputs. From Zh$\rightarrow$En and De$\rightarrow$En, \method outperforms all unsupervised metrics significantly. In comparison to supervised metrics, \method surpasses both BLEURT and COMET in Zh$\rightarrow$En, and achieves a comparable performance to the COMET and 0.016 Kendall correlation gap to BLEURT at De$\rightarrow$En. 

\begin{table}[t]\small
    \centering
    %\begin{tabular}{@{}l|lllllllll@{}}
    \begin{tabular}{@{}l|ccccc@{}}
        \toprule
        \multicolumn{2}{c}{\multirow{2}{*}{\bf Model Name}} & \multicolumn{4}{c}{\bf Machine Translation (WMT21)} \\
        \cmidrule(r){3-6}

        \multicolumn{2}{c}{} & News & TED & Overall & $\Delta$ \\
        \midrule
        \parbox[t]{5mm}{\multirow{2}{*}{\rotatebox[origin=c]{90}{\shortstack{With}}}}
        & \multicolumn{1}{c}{BLEURT} & 0.305 & 0.243  & 0.274 & 0.062\\
        & \multicolumn{1}{c}{COMET(DA)} & 0.300 & 0.240 & 0.270 & 0.060\\
        \cmidrule(l){1-6}

        \parbox[t]{5mm}{\multirow{7}{*}{\rotatebox[origin=c]{90}{\shortstack{W.o Supervision}}}}
        & \multicolumn{1}{c}{TER} & 0.154 & 0.134  & 0.144 & 0.020\\
        & \multicolumn{1}{c}{BLEU} & 0.130 & 0.103 & 0.117 & 0.027\\
        & \multicolumn{1}{c}{ChrF} & 0.158 & 0.135  & 0.147 & 0.023\\
        & \multicolumn{1}{c}{BARTScore} & 0.140 & 0.111 & 0.126 & 0.029\\
        & \multicolumn{1}{c}{BERTScore} & 0.232 & 0.194 & 0.213 & 0.038\\
        & \multicolumn{1}{c}{PRISM} & 0.239 & 0.216  & 0.228 & 0.023\\
        & \multicolumn{1}{c}{SEScore} & 0.273 & 0.235  & 0.254 & 0.038\\
        & \multicolumn{1}{c}{\method} & \textbf{0.287} & \textbf{0.265} &  \textbf{0.276} & 0.022\\
        \bottomrule
    \end{tabular}
    \caption{Segment-level Kendall correlation for WMT21 (En-De and Zh-En) News and TED Testing sets. $\Delta$ indicates the absolute correlation difference between News and TED. Overall indicates the metrics' average performance of News and TED domains. \method outperforms all baseline metrics at TED domain significantly.}
    \label{tab:mt-abalation}
\end{table}

\subsection{\method achieves consistent superior performance across different domains}
\label{sec:domain}
We investigate the domain influence on the evaluation metrics when shifting the testing set from News to TED. As shown in Table \ref{tab:mt-abalation}, all metrics have lower Kendall correlations in TED compared to those in News. We conjectured that the cause is due to the domain differences between the two testing suites. Unlike News, TED contains sentences with informal and disfluent language styles \cite{freitag-etal-2021-results}. 
The supervised learned metrics have the largest gap when shifting the domain from News to TED. The reason is that the entire supervised human rating data is from the News domain only. Although the rule-based metrics (TER. BLEU and Chrf) have relatively lower overall correlations, their correlation is less influenced by the domain shift, with an average 0.023 Kendall correlation difference. Unsupervised learned metrics can be less influenced by domain shift compared to supervised metrics. However, they still have more Kendall correlation drops compared to the rule-based metrics, with an average Kendall correlation 0.032. Most importantly, we observed that \method achieves the highest overall Kendall correlation and achieves the lowest gap (0.022) among learned metrics when shifting between domains. In Section \ref{sec:abalation-quantity}, we demonstrate that \method can take advantage of the data scale and improve its performance on TED while scaling up the data. Full results can be found in Appendix Table \ref{tab:full-lang-results}.
\begin{figure*}[t!]
    \centering
    \begin{subfigure}{.32\textwidth}
    \includegraphics[width=1\textwidth]{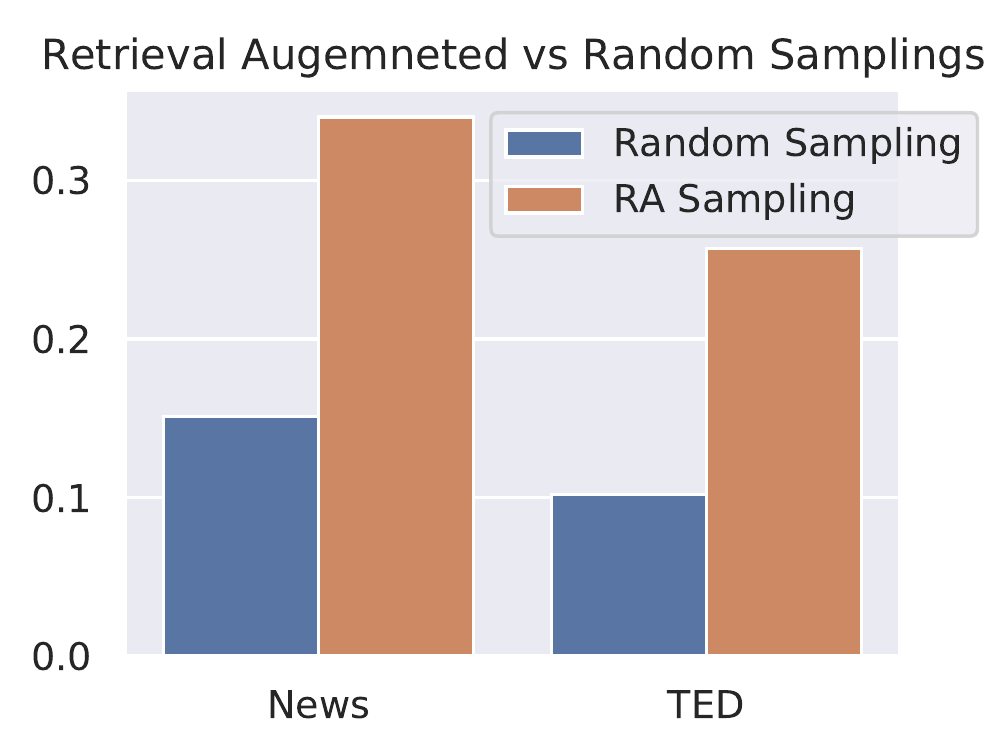}
    \end{subfigure}
    \begin{subfigure}{.32\textwidth}
    \includegraphics[width=1\textwidth]{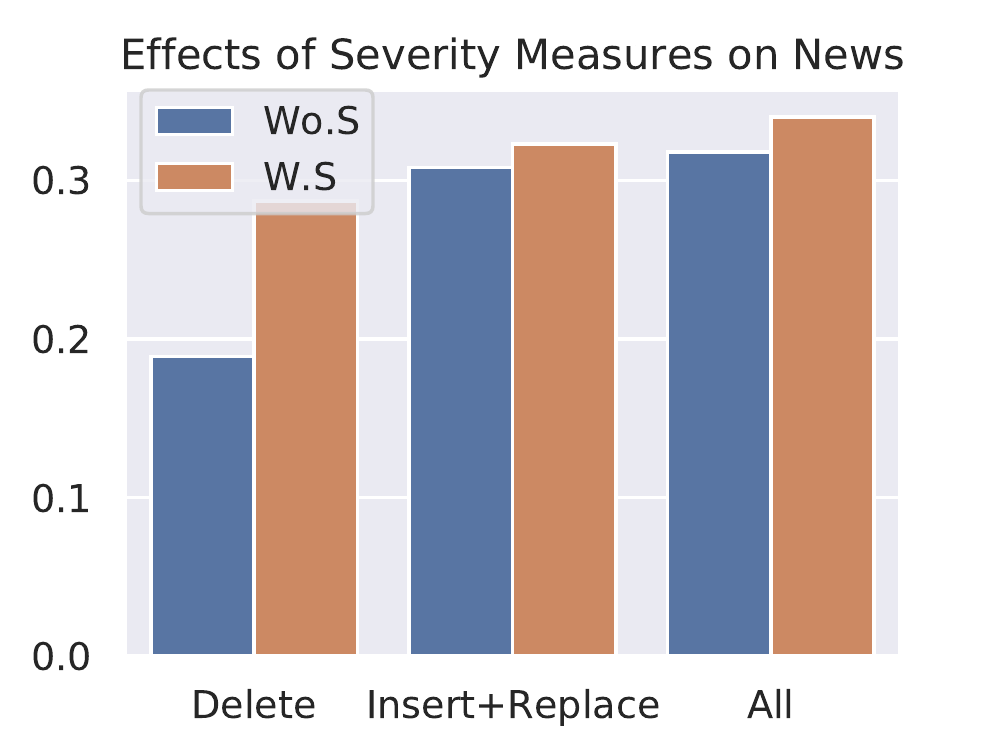}
    \end{subfigure}
    \begin{subfigure}{.32\textwidth}
    \includegraphics[width=1\textwidth]{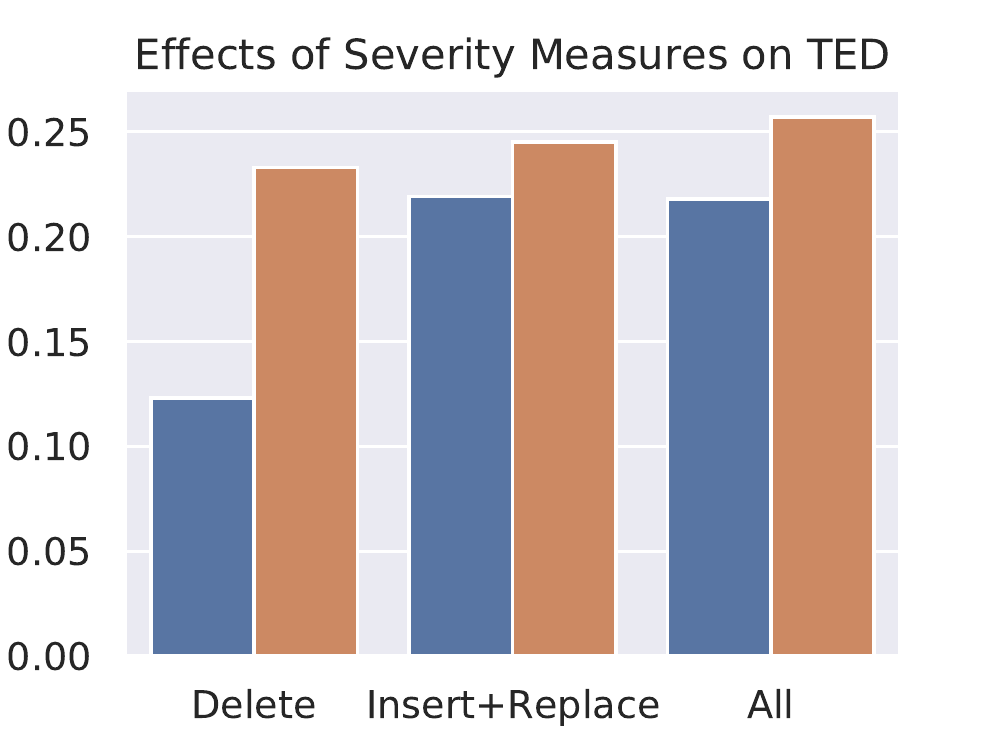}
    \end{subfigure}
    \caption{Left figure indicates the comparisons between \method trained from retrieval augmented synthesis and random token transformations. Middle and right figure indicate individual operations contribute to final \method and effects of severity measures at News and TED domains. W.S means with severity measures and Wo.S means without severity measures.}
    \label{fig:funct_type}
\end{figure*}

\subsection{Quantitative Analysis}
To validate the ideas in \method, we investigate the impact of data scaling, the risks of human rating supervision, effects of our retrieval augmented (RA) synthesis, ablation on RA operations and severity measures, and interpretation of \method at different evaluation dimensions. We include the effects of model initialization in Appendix \ref{sec:model_init}.

% \subsubsection{W and W.o Human Outputs}
% First, we contrast metrics in with and without human output settings at News testing set. All metrics have better performance in only ranking system outputs and on average 1\% performance drops when ranking both model and human outputs. \method outperforms all rule-based metrics with a large margin. As a unsupervised learned metric, \method outperforms all top performing unsupervised learned metrics in all settings at News domain, with the average 7\% absolute Kendall correlation improvement. Surprisingly, compared to the supervised learned metrics, COMET and BLEURT, \method has only average 0.85\% and 1.1\% Kendall correlation gap to them in with and without human output settings. 

\subsubsection{Law of the Data Scaling}
\label{sec:abalation-quantity}
% Change the writing direction: from just Kendall correlation but on error distributions
We study the scaling effects on \method's performance, by testing checkpoints trained at 0.5M, 1M, 2M, and 4M training samples. For both Zh-En and En-De across two domains, we observe the sharp performance improvements at first 1M pretraining data. Larger pretraining data quantity leads to higher human correlations for both language directions. We get 2.5\% and 1.8\% improvements in Zh-En, and 2.5\% and 1.1\% improvements in En-De at 2M and 3M data points. Performance saturates from 3M to 4M synthetic data, with around 0.5\% improvements in both language directions. This suggests that a data scale of 1M can train a competitive metric, and larger data can gradually improve the metric to fit into a general domain.

% We observe that larger pretraining data quantity can overall lead to higher human correlations for both language directions. For Zh-En, we obtain 2.5\% and 1.8\% Kendall correlation improvements at 2M and 3M data points respectively. For En-De, we obtain 2.5\% and 1.1\% Kendall correlation improvements at 2M and 3M data points respectively. The performance starts to saturated from 3M to 4M synthetic data (around 0.5\% improvements in both language directions). This result suggests that data scale around 1M can train a metric with competitive performance and larger pretraining data can gradually improve pretrained metric to fit into a general domain.

% \begin{table}
% \resizebox{\textwidth}{!}{% 
% \capbtabbox{
%         \resizebox{0.95\textwidth}{!}{%  
%         \begin{tabular}{@{}lllll@{}}
%             \toprule \multicolumn{1}{c}{} & \multicolumn{2}{c}{\bf En$\rightarrow$De} & \multicolumn{2}{c}{\bf Zh$\rightarrow$En} \\
%             \cmidrule(r){2-3}
%             \cmidrule(l){4-5}
%             \multicolumn{1}{c}{Training Settings} & News & TED & News & TED \\ \midrule
%             \multicolumn{1}{c}{Unsupervised} & 0.227 & 0.258 & 0.347
%             & 0.271 \\
%             \multicolumn{1}{c}{Supervised (News)} & 0.254 & 0.269 & 0.369 & 0.188 \\
%             \bottomrule
%         \end{tabular}
%         }
% }{
%  \caption{\method's segment-level Kendall correlation under the unsupervised and supervised settings at News and TED testing sets}
%  \label{tab:test_data_stats}
% }
% }
% \end{table}

\begin{table}[ht]
%\resizebox{0.90\textwidth}{!}{% 
%\capbtabbox{
        \resizebox{\textwidth}{!}{%  
        \begin{tabular}{@{}lllll@{}}
            \toprule \multicolumn{1}{c}{} & \multicolumn{4}{c}{\bf Machine Translation WMT21} \\
            \cmidrule(r){2-5}

            \multicolumn{1}{c}{Metric Name} & News & TED & Overall & $\Delta$\\ \midrule
            \multicolumn{1}{c}{COMET} & 0.300 & 0.240 & 0.270 & 0.060\\
            \multicolumn{1}{c}{BLEURT} & 0.305 & 0.243 & 0.274 & 0.062\\
            \multicolumn{1}{c}{\method+FT} & \textbf{0.312} & 0.229 & 0.271 & 0.083 \\
            \multicolumn{1}{c}{\method} & 0.287 & \textbf{0.265} & \textbf{0.276}
            & 0.022 \\
            \bottomrule
        \end{tabular}
        }
%}{
 \caption{Segment-level Kendall correlation under the \method and fine-tuned (FT) \method with supervised COMET and BLEURT at WMT21 News and TED testing sets. Overall measures the overall correlation of two domains and $\Delta$ indicates the correlation gap between two domains.}
 \label{tab:fine-tune}
%}
%}
\end{table}

\subsubsection{Danger of Fine-tuning}
\label{sec:abalation-supervised}
To address the question "Can fine-tuning always lead to better performance?", we fine-tune \method on existing 350K English and 59K German WMT17-19 News domain human rating data \footnote{Like COMET, we use WMT17-19 DA human ratings. BLEURT uses WMT15-19 DA results for its training dataset.}. For each domain, we report the average Kendall correlation between En-De and Zh-En. In Table \ref{tab:fine-tune}, \method+FT improves by 8.7\% over \method in the News domain. Additionally, it outperforms both BLEURT and COMET in the News domain with 4\% and 2.3\% Kendall improvements respectively. Despite the improvements in News, the supervised \method has a 13.6\% correlation drop in the TED testing set, resulting in a larger correlation gap (0.083) between News and TED domains. This confirms our assumption that fine-tuning with domain-specific human ratings can fit the metric tightly to the trained domain distribution, but may decrease its generalization ability across domains. The unsupervised \method achieves the highest overall Kendall correlation across two domains. 

\subsubsection{Effectiveness of Retrieval Augmented Synthesis} 
\label{sec:retrieval_random}
In Figure \ref{fig:funct_type}, we demonstrate the superior performance of retrieval augmented (RA) data construction compared to the random token transformations. Our observation is that most of the random in-batch tokens have low co-occurrence probabilities with their contexts. The sentence embeddings from those text transformations can be easily distinguished from the anchor embedding, by the pretrained language model \cite{https://doi.org/10.48550/arxiv.1911.02116}. Therefore, further pretraining on negative samples with random token transformations does not lead to significant correlation improvements. We empirically demonstrate that RA data construction improves random token insert/replace by 114\% in the News and TED domains.  

\subsubsection{Ablation on RA operations}
\label{sec:abalation_ra}
To evaluate the performance of each component at our retrieval augmented synthesis, we separately trained checkpoints with synthetic data that 1) contains delete operations only; 2) contains insert and replace operations according to our scheduled locations; 3) contains all operations with scheduled positions. To exclude the effects from the severity measures, we do not assign severity measures for each error and instead label each sentence with the number of errors it contains. In Figure \ref{fig:funct_type}, we observe that our RA insert/replace contributes the most of the human correlations, 0.308 at News and 0.219 at TED. This suggests that our scheduled positions to insert and replace are important to construct realistic synthetic sentences and learning meaningful embeddings. Despite the simple scheme, delete-only construction can achieve competitive performance, with Kendall correlation 0.189 and 0.123 in News and TED respectively. By combining all operations, the aggregated effect can further improve the Kendall correlation 3.2\% at News. 

\subsubsection{Effects of Severity Measures} 
\label{sec:abalation_se}
In Figure \ref{fig:funct_type}, we empirically verify two of our severity measures: 1) IDF-based 2) XLM-based approaches. Our IDF-based severity measures on delete operation can improve 51.9\% Kendall correlation at News and 81.3\% at TED. Our XLM-based severity measures on insert and delete can improve 4.9\% at News and 11.9\% at TED. Lastly, the joint effect of two severity measures can improve \method without severity measures by 6.92\% at News and 17.9\% at TED. 

\begin{figure}
    \centering
    \includegraphics[width=0.95\linewidth]{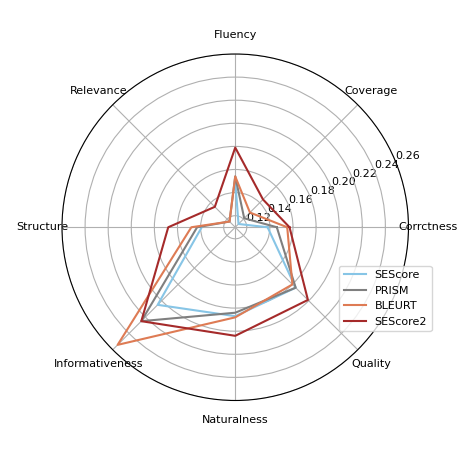}
    \caption{Kendall correlations at Multi-dimensional WebNLG and BAGEL benchmarks. We select top four performing metrics.}
    \label{fig:sescore2_aspects}
\end{figure}

\section{How to interpret \method?}
In order to interpret the evaluation dimensions of \method, we conducted multi-dimensional human correlations in WebNLG and BAGEL benchmarks. In Figure \ref{fig:sescore2_aspects}, we observe that \method achieves the highest Kendall correlation across all dimensions except informativeness compared to the BLEURT. At WebNLG and BAGEL, \method is most correlated to fluency, text structure, naturalness, and quality, which leads the second highest metric BLEURT by 16.2\% and 13.5\%, 8.5\% and 10.6\% respectively. To conclude, despite producing an overall score, \method can be a great indicator of diverse evaluation aspects of text generation. In particular, \method has a significant advantage over baseline metrics in terms of the quality and fluency aspects of quality assessment. Full results are in Appendix Table \ref{tab:webnlg-multi} and Table \ref{tab:dialogue-multi}. We further rescaled our results to the predefined range ($0$ to $-50$) to ensure consistency and interpretability across domains and tasks (See Appendix Section \ref{sec:score_range} for implementation details). 

\section{Supported Languages}
Currently, \method supports English, German, Japanese, Spanish, Chinese and Russian. Our pipeline in extending \method to future languages is generic. It is straightforward to extend \method to up to 100 languages i.e. any language supported in XLM.  
%(Afrikaans, Albanian, Amharic, Arabic, Armenian, Assamese, Azerbaijani, Basque, Belarusian, Bengali, Bengali Romanized, Bosnian, Breton, Bulgarian, Burmese, Burmese, Catalan, Chinese (Simplified), Chinese (Traditional), Croatian, Czech, Danish, Dutch, English, Esperanto, Estonian, Filipino, Finnish, French, Galician, Georgian, German, Greek, Gujarati, Hausa, Hebrew, Hindi, Hindi Romanized, Hungarian, Icelandic, Indonesian, Irish, Italian, Japanese, Javanese, Kannada, Kazakh, Khmer, Korean, Kurdish (Kurmanji), Kyrgyz, Lao, Latin, Latvian, Lithuanian, Macedonian, Malagasy, Malay, Malayalam, Marathi, Mongolian, Nepali, Norwegian, Oriya, Oromo, Pashto, Persian, Polish, Portuguese, Punjabi, Romanian, Russian, Sanskri, Scottish, Gaelic, Serbian, Sindhi, Sinhala, Slovak, Slovenian, Somali, Spanish, Sundanese, Swahili, Swedish, Tamil, Tamil Romanized, Telugu, Telugu Romanized, Thai, Turkish, Ukrainian, Urdu, Urdu Romanized, Uyghur, Uzbek, Vietnamese, Welsh, Western, Frisian, Xhosa, Yiddish). 
To obtain reliable severity measures, we currently support 14 languages (Ar, Bg, De, El, Es, Fr, Hi, Ru, Sw Th, Tr, Ur, Vi and Zh). For detailed limitation discussion about severity measures, please refer to Section \ref{sec:limit}.

%% file: 050conclusion.tex
We propose a novel retrieval augmented synthesis method for generating diverse and realistic errors on a large scale, with varying severity levels. Our experiment demonstrates, \method, with its human-aligned self-supervised objective, can outperform prior metrics or match supervised metrics in four text generation tasks across three languages. Lastly, we demonstrate \method can correlate well with multiple evaluation aspects, such as fluency and quality.

%% file: 060limitation.tex
One potential improvement to this work is the development of a method to evaluate the accuracy of the severity measure component. We have demonstrated the effectiveness of \method with a severity measure through improved Kendall correlations for various types of retrieval augmented synthesis in Figure \ref{fig:funct_type}. However, there is currently no widely accepted way to quantitatively measure the accuracy of the severity labels. This is because there is no existing dataset that can be used to benchmark severity measures. While \citet{Freitag_2021, freitag-etal-2021-results} have released MQM annotations with error spans for each segment, these annotations often include compositional errors that prevent the evaluation of individual severity labels without also considering other errors in the sentence. A potential future direction for this research would be to create a benchmark dataset that would allow direct assessment of individual severity estimation or explore alternative methods for evaluating the accuracy of severity measures.

Second, we have not been able to test \method on low-resource languages due to the lack of MQM-annotated testing sets in these languages. However, we have demonstrated that \method can still perform well without severity estimation by outperforming top unsupervised metrics such as BERTScore, BARTScore and PRISM as shown in Figure \ref{fig:funct_type}. This suggests that \method may be useful for low-resource languages since parallel corpora are not available for most low-resource language settings. To further verify this, a potential future direction would be to create testing sets with MQM labels for low-resource languages, to test the performance of \method and other learned metrics in such scenarios.

% Second limitation of this work is that we are not able to verify \method over low resource language. Due to the scarcity of low resource language experts, there is no testing set available for low resource language with MQM annotations. To simulate a low resource language setting, we should assume there is no parallel corpora for language $X$, so that severity estimation is not possible. In Figure \ref{fig:funct_type}, we demonstrate that \method without severity estimation can still outperform top performing unsupervised metrics like BERTScore, BARTScore and PRISM. However, we can not exclude the performance gains which come from our backbone (ex. RemBert). A potential future direction would be to create a testing set on low resource language with MQM labels. 

Lastly, since \method is based on proximity between reference and model output, its capabilities for open-ended text generation tasks have not yet been fully explored. This presents an opportunity for future research to investigate the potential of this method in such scenarios.

%% file: 070impact.tex
We hired three human raters to annotate WMT21 De$\rightarrow$En testing set. The evaluated text is publicly available machine translation testing set which contains no sensitive or explicit languages. There is no risk of exposing annotators' identities and they are fully aware the usage of their collected dataset. We use the standard MQM human evaluation procedures \cite{Freitag_2021} and all annotaters are experienced with this evaluation protocol. All collected human annotations will be released at the camera ready. The hourly salary for the raters are well above the minimum wage in the local region. Details can be found in Appendix \ref{sec:human_annotations}.

%% file: 080appendix.tex
\begin{table}
\resizebox{\textwidth}{!}{%  
\begin{floatrow}
\capbtabbox{
        \resizebox{0.95\textwidth}{!}{%
        \begin{tabular}{@{}llllll@{}}
            \toprule
            \multicolumn{6}{c}{\bf WebNLG Data-to-Text Generation} \\
            \cmidrule(r){1-6}
            \multicolumn{1}{c}{Model Name} & Cor & Cov & Flu  & Rel & Str\\ \midrule
        %     \multicolumn{1}{c}{ROUGE-1} & 0.178* & 0.135* & 0.103*\\
        %     \multicolumn{1}{c}{ROGUE-2} & 0.151* & 0.134* & 0.109*\\
        %     \multicolumn{1}{c}{ROGUE-L} & 0.146* & 0.126* & 0.107* \\
        %   \multicolumn{1}{c}{MoverScore} & 0.209* & 0.140* & 0.114*\\
        \multicolumn{1}{c}{TER} & -0.075* & -0.060* & -0.082* & -0.067* & -0.082*\\
        \multicolumn{1}{c}{BLEU} & 0.077* & 0.062* & 0.075* & 0.065* & 0.070* \\
        \multicolumn{1}{c}{ChrF} & 0.088* & 0.087* & 0.082* & 0.076* & 0.073* \\
        \multicolumn{1}{c}{BARTScore} & 0.096* & 0.085* & 0.107* & 0.079* & 0.102* \\
        \multicolumn{1}{c}{BERTScore} & 0.141* & 0.110* & 0.143* & 0.108* & 0.142* \\
        \multicolumn{1}{c}{SEScore} & 0.138* & 0.114* & 0.150* & 0.108* & 0.139* \\
        \multicolumn{1}{c}{PRISM} & 0.146* & 0.121* & 0.154* & 0.117* & 0.143*\\
        \multicolumn{1}{c}{BLEURT} & 0.155 & 0.128* & 0.154* & 0.117* & 0.148*\\
        \midrule
        \multicolumn{1}{c}{\method} & \textbf{0.157} & \textbf{0.144} & \textbf{0.179} & \textbf{0.135} & \textbf{0.168}\\
        \bottomrule
        \end{tabular}
    }
}{
 \caption{Segment-level Kendall Correlation on WebNLG Data-to-Text generation. * indicates that \method significantly outperforms the baseline metric (p<0.05). Cor, Cov, Flu, Rel and Str represents Correctness, Coverage, Fluency, Relevance and Text Structure respectively.}
 \label{tab:webnlg-multi}
}
\end{floatrow}
}
\end{table}

\begin{table}
\resizebox{\textwidth}{!}{%  
\begin{floatrow}
\capbtabbox{
        \resizebox{0.95\textwidth}{!}{%
        \begin{tabular}{@{}llllll@{}}
            \toprule
            \multicolumn{6}{c}{\bf WebNLG Data-to-Text Generation} \\
            \cmidrule(r){1-6}
            \multicolumn{1}{c}{Model Name} & Cor & Cov & Flu  & Rel & Str\\ \midrule
        %     \multicolumn{1}{c}{ROUGE-1} & 0.178* & 0.135* & 0.103*\\
        %     \multicolumn{1}{c}{ROGUE-2} & 0.151* & 0.134* & 0.109*\\
        %     \multicolumn{1}{c}{ROGUE-L} & 0.146* & 0.126* & 0.107* \\
        %   \multicolumn{1}{c}{MoverScore} & 0.209* & 0.140* & 0.114*\\
        \multicolumn{1}{c}{TER} & -0.109* & -0.087* & -0.12* & -0.067* & -0.082*\\
        \multicolumn{1}{c}{BLEU} & 0.112* & 0.091* & 0.110* & 0.095* & 0.102* \\
        \multicolumn{1}{c}{ChrF} & 0.129* & 0.127* & 0.121* & 0.112* & 0.107* \\
        \multicolumn{1}{c}{BARTScore} & 0.142* & 0.124* & 0.158* & 0.115* & 0.151* \\
        \multicolumn{1}{c}{BERTScore} & 0.208* & 0.161* & 0.211* & 0.158* & 0.210* \\
        \multicolumn{1}{c}{SEScore} & 0.202* & 0.167* & 0.220* & 0.158* & 0.204* \\
        \multicolumn{1}{c}{PRISM} & 0.214* & 0.176* & 0.227* & 0.169* & 0.211*\\
        \multicolumn{1}{c}{BLEURT} & 0.225 & 0.186* & 0.226* & 0.170* & 0.217*\\
        \midrule
        \multicolumn{1}{c}{\method} & \textbf{0.231} & \textbf{0.212} & \textbf{0.263} & \textbf{0.197} & \textbf{0.248}\\
        \bottomrule
        \end{tabular}
    }
}{
 \caption{Segment-level Spearman Correlation on WebNLG Data-to-Text generation. * indicates that \method significantly outperforms the baseline metric (p<0.05). Cor, Cov, Flu, Rel and Str represents Correctness, Coverage, Fluency, Relevance and Text Structure respectively.}
 \label{tab:webnlg-spearman}
}
\end{floatrow}
}
\end{table}

\section{Pretraining Data Collection}
\label{sec:pretrain_data_construct}
For Chinese-to-English, we collect 20M sentence pairs from UN Parallel \cite{ziemski-etal-2016-united}, News Commentary \cite{tiedemann-2012-parallel} and CWMT corpus \cite{barrault-etal-2019-findings}. For English-to-German, we collect 4.5M sentence pairs from Europarl \cite{koehn-2005-europarl}, Common Crawl \cite{K_dela_2017} and News Commentary. For English-to-Japanese, We collect 18M sentence pairs from News Complimentary, WikiMatrix \cite{schwenk-etal-2021-wikimatrix} and En-Ja subtitle corpus \cite{pryzant-etal-2018-jesc}.

\begin{table}
\resizebox{\textwidth}{!}{% 
\capbtabbox{
        \resizebox{0.95\textwidth}{!}{%  
        \begin{tabular}{@{}lllllll@{}}
            \toprule \multicolumn{1}{c}{} & \multicolumn{3}{c}{\bf News} & \multicolumn{3}{c}{\bf TED} \\
            \cmidrule(r){2-4}
            \cmidrule(l){5-7}
            \multicolumn{1}{c}{LP} & \#H & \#Sys & \#Sents & \#H & \#Sys & \#Sents \\ \midrule
            \multicolumn{1}{c}{Zh$\rightarrow$En} & 2 & 13 & 650 & 1 & 13 & 529\\
            \multicolumn{1}{c}{En$\rightarrow$De} & 4 & 13 & 527 & 1 & 14 & 529\\
            \multicolumn{1}{c}{De$\rightarrow$En} & 1 & 9 & 100 & - & - & -\\
            \bottomrule
        \end{tabular}
        }
}{
 \caption{Human annotation statistics for Machine Translation (MT) task. \#H refers to the number of humans, \#Sys refers to the number of MT systems and \#Sents refers to the number of annotated samples per system.}
 \label{tab:test_data_stats}
}
}
\end{table}

\section{Index Table Construction}
\label{sec:index_table}
We use LASER library\footnote{https://github.com/facebookresearch/LASER} to compute all the sentence embeddings and use Faiss library \footnote{https://github.com/facebookresearch/faiss} to build the index table for English, German and Japanese. We used 8*A100 GPUs for the index table construction. The duration for building index table for English, German and Japanese is 48 hours, 24 hours and 48 hours respectively. From the constructed index table, we extracted 128 nearest neighbors for each text sentence. To ensure our learned metrics can cover diverse domains and tasks, we sample millions of raw sentences from diverse domains of corpuses and build ten-million scale index tables. Detailed statistics are discussed at Section \ref{sec:pretraining_data_stats}.

\section{Margin-based Criterion}
\label{sec:laser_setting}
We follow the implementation of margin criterion in \cite{schwenk-etal-2021-wikimatrix}. We set the threshold of margin criterion to be 1.06 and  extract 128 nearest neighbors to estimate mutual translation capability.

\section{Pretrained Model Initialization}
\label{sec:model_init}
All checkpoints from Rembert are trained for 15,000 iterations and all checkpoints from XLM-R are trained for 30,000 iterations.
Since our pipeline utilized pretrained model, we try to answer the question that with the setting, can different pretrained model initialization lead to different performance? In particular, we studied two prior used pretrained models: RemBERT (used by BLEURT) and XLM-R (used by COMET). 
% RemBERT is a larger pretrained language model 32-layer versus 24-layer used in XLM-R. 
Based on prior study \cite{chung2020rethinking}, RemBERT emperically outperforms XLM-R over multiple multilingual downstream tasks. In Table \ref{tab:init}, we demonstrates that compared to XLM-R initialization, our \method with RemBERT initialization can further improve Kendall correlations in all language directions. This finding suggests that \method with a better pretrained model initialization can increase its learning capacity of score distribution and improve its correlations to human ratings. 

\begin{table}
\resizebox{\textwidth}{!}{% 
\capbtabbox{
        \resizebox{0.95\textwidth}{!}{%  
        \begin{tabular}{@{}lllll@{}}
            \toprule 
            \multicolumn{1}{c}{} & \multicolumn{2}{c}{\bf Zh$\rightarrow$En} & \multicolumn{2}{c}{\bf En$\rightarrow$De} \\
            \cmidrule(r){2-3}
            \cmidrule(l){4-5}
            \multicolumn{1}{c}{Initialization} & News & TED & News & TED \\ \midrule
            \multicolumn{1}{c}{\method(XLM-R)} & 0.340 & 0.257 & 0.206 & 0.249 \\
            \multicolumn{1}{c}{\method(RemBERT)} & \textbf{0.348} & \textbf{0.271} & \textbf{0.227} & \textbf{0.258} \\
            \bottomrule
        \end{tabular}
        }
}{
 \caption{Segment-level Kendall correlation using different pretrained model initialization on WMT21 En-De and Zh-En at both News and TED domains.}
 \label{tab:init}
}
}
\end{table}

% To understand the failure cases of \method, we conduct detailed analysis of \method at social domain of Zh$\rightarrow$En, chat domain of En$\rightarrow$De and E-commerce domain of En$\rightarrow$De in Section \ref{sec:err_case_study}.

\section{Baseline Implementations}
\label{sec:baseline_implmentations}
For WMT21 News and TED, we use WMT officially released output results \cite{freitag-etal-2021-results} from their official script\footnote{https://github.com/google-research/mt-metrics-eval}. 
We use HuggingFace evaluate module (Open sourced library) to get baseline outputs for BLEURT \cite{sellam-etal-2020-bleurt}, COMET \cite{rei-etal-2020-comet}, SEScore \cite{xu-etal-2022-not}, BLEU \cite{papineni-etal-2002-bleu}, TER \cite{snover-etal-2006-study} and ChrF \cite{popovic-2015-chrf}. For BERTScore \cite{https://doi.org/10.48550/arxiv.1904.09675}, BARTScore \cite{https://doi.org/10.48550/arxiv.2106.11520} and PRISM\cite{thompson-post-2020-automatic}, we use their open sourced Github repository. Specifically, following their recommendations, we use Roberta-large backbone \cite{Liu2019RoBERTaAR} for English assessment of BERTScore. We use multilingual BERTScore \cite{devlin-etal-2019-bert} to assess German and Japanese testing sets. For BARTScore, we use recommended Bart-large-cnn backbone for English testing sets and MBART backbone \cite{liu-etal-2020-multilingual-denoising} for German and Japanese testing sets. For SEScore, we use stratified error synthesis process \cite{xu-etal-2022-not} to construct 120,000 Japanese synthesized texts with pseudo labels and trained a Japanese SEScore. We use this Japanese SEScore to test Japanese testing set.

\section{Human Annotation Procedure}
\label{sec:human_annotations}
We conduct a human evaluation on WMT21 de-en testing set at News domain. We randomly select $10$ systems outputs out of $20$ participating systems. We annotate $100$ testing segments for each selected system. In total, $1000$ testing sentences are annotated. Following the prior study \citet{Freitag_2021}, we hired 3 bilingual linguists in English and German. Following MQM evaluation procedure \cite{Freitag_2021}, each rater can only access the source (in German) and model output (in English), without any reference text. Rater is given all the choices of possible error typologies and definitions of the severity levels. We directly use the instruction of MQM hierarchy, MQM severity levels and MQM annotator guidelines from prior work \cite{Freitag_2021} to our human raters (Please refer to Table 10, Table 11 and Table 12 for specific references). All three raters are well-trained to perform human ratings. The hourly rate for all raters is $70$ Chinese Yuan per hour. The local minimum wage is $23$ Chinese Yuan per hour. Our human evaluation study has no risk of exposing any annotator's identities and text contains neither sensitive or explicit language.

\begin{table}
\resizebox{\textwidth}{!}{%  
\begin{floatrow}
\capbtabbox{
        \resizebox{\textwidth}{!}{%
        \begin{tabular}{@{}llll@{}}
            \toprule
            \multicolumn{4}{c}{\bf BAGEL Dialogue Generation} \\
            \cmidrule(r){1-4}
            \multicolumn{1}{c}{Model Name} & Informativeness & Naturalness & Quality\\ \midrule
        %     \multicolumn{1}{c}{ROUGE-1} & 0.178* & 0.135* & 0.103*\\
        %     \multicolumn{1}{c}{ROGUE-2} & 0.151* & 0.134* & 0.109*\\
        %     \multicolumn{1}{c}{ROGUE-L} & 0.146* & 0.126* & 0.107* \\
        %   \multicolumn{1}{c}{MoverScore} & 0.209* & 0.140* & 0.114*\\
            \multicolumn{1}{c}{TER} & -0.055* & -0.127* & -0.079* \\
            \multicolumn{1}{c}{chrF} & 0.182* & 0.078* & 0.064* \\
            \multicolumn{1}{c}{BLEU} & 0.138* & 0.104* & 0.085* \\
            \multicolumn{1}{c}{BERTScore} & 0.217* & 0.114* & 0.159* \\
            \multicolumn{1}{c}{BARTScore} & 0.183* & 0.114* & 0.183*\\
            \multicolumn{1}{c}{SEScore} & 0.205* & 0.187* & 0.184* \\
            \multicolumn{1}{c}{PRISM} & 0.225 & 0.184* & 0.184*\\
            \multicolumn{1}{c}{BLEURT} & \textbf{0.254} & 0.188* & 0.180*\\
            \midrule
            \multicolumn{1}{c}{\method} & 0.225 & \textbf{0.204} & \textbf{0.199}\\
            \bottomrule
        \end{tabular}
        }
}{
 \caption{\footnotesize Segment-level Kendall Correlation on BAGEL dialogue generation. * indicates that \method significantly outperforms the baseline metric (p<0.05).}
 \label{tab:dialogue-multi}
}
\end{floatrow}
}
\end{table}

\begin{table}
\resizebox{\textwidth}{!}{%  
\begin{floatrow}
\capbtabbox{
        \resizebox{\textwidth}{!}{%
        \begin{tabular}{@{}llll@{}}
            \toprule
            \multicolumn{4}{c}{\bf BAGEL Dialogue Generation} \\
            \cmidrule(r){1-4}
            \multicolumn{1}{c}{Model Name} & Informativeness & Naturalness & Quality\\ \midrule
        %     \multicolumn{1}{c}{ROUGE-1} & 0.178* & 0.135* & 0.103*\\
        %     \multicolumn{1}{c}{ROGUE-2} & 0.151* & 0.134* & 0.109*\\
        %     \multicolumn{1}{c}{ROGUE-L} & 0.146* & 0.126* & 0.107* \\
        %   \multicolumn{1}{c}{MoverScore} & 0.209* & 0.140* & 0.114*\\
            \multicolumn{1}{c}{TER} & -0.073* & -0.170* & -0.105* \\
            \multicolumn{1}{c}{chrF} & 0.244* & 0.103* & 0.083* \\
            \multicolumn{1}{c}{BLEU} & 0.182* & 0.138* & 0.114* \\
            \multicolumn{1}{c}{BERTScore} & 0.289* & 0.155* & 0.212* \\
            \multicolumn{1}{c}{BARTScore} & 0.247* & 0.155* & 0.247*\\
            \multicolumn{1}{c}{SEScore} & 0.272* & 0.248* & 0.243* \\
            \multicolumn{1}{c}{PRISM} & 0.305 & 0.248* & 0.247*\\
            \multicolumn{1}{c}{BLEURT} & \textbf{0.343} & 0.252* & 0.241*\\
            \midrule
            \multicolumn{1}{c}{\method} & 0.304 & \textbf{0.273} & \textbf{0.264}\\
            \bottomrule
        \end{tabular}
        }
}{
 \caption{\footnotesize Segment-level Spearman Correlation on BAGEL dialogue generation. * indicates that \method significantly outperforms the baseline metric (p<0.05).}
 \label{tab:dialogue-multi-spearman}
}
\end{floatrow}
}
\end{table}

\begin{table}
\resizebox{0.5\linewidth}{!}{%  
\begin{tabular}{@{}lll@{}}
    \toprule \multicolumn{1}{c}{} & \multicolumn{2}{c}{\bf WMT20} \\
    \cmidrule(r){2-3}
    \multicolumn{1}{c}{$\gamma$} & Zh$\rightarrow$En & En$\rightarrow$De \\ \midrule
    \multicolumn{1}{c}{0.1} & 0.160 & 0.288 \\
    \multicolumn{1}{c}{0.2} & 0.156 & 0.283 \\
    \multicolumn{1}{c}{0.3} & 0.158 & 0.286 \\
    \multicolumn{1}{c}{0.4} & 0.154 & 0.281 \\
    \bottomrule
\end{tabular}
}
{
 \caption{Greedy search $\gamma$ from $0.1$ to $0.4$ and construct label based on each threshold. We use the triples constructed from different $(x,y,s)$ to test \method's Kendall correlation on WMT20 En-De and Zh-En}
 \label{tab:threshold_search}
}
\end{table}

\section{Hyperparameters on Severity Measures}
\label{sec:choices_threshold}
We conducted preliminary experiments over hyperparameter choices for the severity measures. We hand-crafted 50 severity examples (each example only contains one major or minor example) to select the $\gamma$ and $\lambda$. We determine the range of $\gamma$ to be from 0.1 to 0.4 and $\lambda$ to be 1. To select the best $\gamma$, we use our retrieval augmented perturbation to generate 200k German and English synthesized sentences, respectively. We greedy searched threshold $\gamma$ from ${0.1, 0.2, 0.3, 0.4}$, for both languages. From Table \ref{tab:threshold_search}, we construct labels based on each possible $\gamma$ and construct corresponding training triples $(x,y,s)$. Therefore, we obtain four checkpoints in each language direction. Based on the Kendall correlations, we concluded that $\gamma$ is not a sensitive hyperparameter. Our best hypothesis is that the tokens that are not likely to occur under the source and target contexts have low probabilities (ex. $0.01$). Therefore, choosing the specific $\gamma$ will not significantly affect the accuracy of the severity measures. In the end, we select $\gamma = 0.1$ for both En-De and Zh-En. 

\section{Score Range Rescaling}
\label{sec:score_range}
In our pretraining data, the synthetic score range is between $0$ to $-50$. However, due to the last activation Tanh in our model, the learned score range is constrained between a positive constant $h$ and a negative constant $l$. As a result, despite having a great ranking capability, our score is not interpretable for users. Ideally, we want our score to lie within the pre-defined range of $0$ to $-50$ and remain invariant across domains and tasks.

To achieve this goal, we collected a large-scale dataset of raw data (2M) from Wikipedia for the target language. We randomly grouped 1M pairs without replacements. Given the random nature of the pairs, we hypothesize that they likely have low semantic or syntactic similarity, thus prohibiting low scores. Consequently, we can obtain a lower score bound $l$ from this set. We calculate score $l$ by averaging \method for all 1M sentence pairs in the set. Furthermore, we randomly select 1M sentences, and for each of them, we feed it twice into the forward layers of \method with a dropout rate of $0.1$. In this case, the two computed embeddings should be nearly identical, allowing us to derive an upper bound $h$ for the score range. We obtain score $h$ by averaging \method for all 1M embedding pairs in this set. 

\begin{equation}
\small
\begin{split}
\method_{Rescale} = (\frac{\method-l}{h-l}-1)*50
\end{split}
\label{eqn:normalize}
\end{equation}

From Eqn \ref{eqn:normalize}, we can normalize our \method roughly between $0$ to $-50$. We can use our pre-defined score for major errors (score: $-5$) and minor errors (score: $-1$) to interpret the final results. By normalizing the score, we can ensure better interpretability and maintain consistency across different domains and tasks.

\begin{table*}[t]\small
    \centering
    \begin{tabular}{@{}l|cccccc@{}}
        \toprule
        \multicolumn{2}{c}{\multirow{2}{*}{\bf Model Name}} & \multicolumn{2}{c}{\bf Zh$\rightarrow$En} & \multicolumn{2}{c}{\bf En$\rightarrow$De} & \multicolumn{1}{c}{\bf De$\rightarrow$En}\\
        \cmidrule(r){3-4}
        \cmidrule(l){5-6}
        \cmidrule(l){7-7}

        \multicolumn{2}{c}{} & News & TED  & News & TED & News\\ \midrule
        \parbox[t]{5mm}{\multirow{2}{*}{\rotatebox[origin=c]{90}{\shortstack{With}}}}
        & \multicolumn{1}{c}{BLEURT} & 0.354 & 0.224 & 0.252 & 0.252 & 0.266\\
        & \multicolumn{1}{c}{COMET(DA)} & 0.360 & 0.220 & 0.239 & 0.259 & 0.250\\
        \cmidrule(l){1-7}
        %& \multicolumn{1}{c}{W/o Human Labels} & Kendall & Pearson & Kendall & Pearson & Kendall & Pearson & Kendall & Pearson\\ \midrule
        \parbox[t]{5mm}{\multirow{7}{*}{\rotatebox[origin=c]{90}{\shortstack{W.o Supervision}}}}
        & \multicolumn{1}{c}{BLEU} & 0.176 & 0.092 & 0.083 & 0.113 & 0.089\\
        & \multicolumn{1}{c}{ChrF} & 0.201 & 0.124 & 0.114 & 0.147 & 0.098\\
        & \multicolumn{1}{c}{TER} & 0.210 & 0.136 & 0.098 & 0.131 & -0.060\\
         & \multicolumn{1}{c}{BERTScore} & 0.296 & 0.199 & 0.169 & 0.199 & 0.205\\
         & \multicolumn{1}{c}{BARTScore} & 0.262 & 0.154 & 0.038 & 0.001 & 0.047\\
         & \multicolumn{1}{c}{PRISM} & 0.285 & 0.194 & 0.192 & 0.238 & 0.174\\
        & \multicolumn{1}{c}{SEScore} & 0.334 & 0.228 & 0.211 & 0.241 & 0.249\\
        & \multicolumn{1}{c}{\method} & \textbf{0.347} & \textbf{0.271} & \textbf{0.227} & \textbf{0.258} & \textbf{0.250}\\
        \bottomrule
    \end{tabular}
    \caption{Segment-level Kendall correlation on En-De and Zh-En for WMT21 News and TED domains. }
    \label{tab:full-lang-results}
\end{table*}

\begin{table*}[t]\small
    \centering
    \begin{tabular}{@{}l|cccccc@{}}
        \toprule
        \multicolumn{2}{c}{\multirow{2}{*}{\bf Model Name}} & \multicolumn{2}{c}{\bf Zh$\rightarrow$En} & \multicolumn{2}{c}{\bf En$\rightarrow$De} & \multicolumn{1}{c}{\bf De$\rightarrow$En}\\
        \cmidrule(r){3-4}
        \cmidrule(l){5-6}
        \cmidrule(l){7-7}

        \multicolumn{2}{c}{} & News & TED  & News & TED & News\\ \midrule
        \parbox[t]{5mm}{\multirow{2}{*}{\rotatebox[origin=c]{90}{\shortstack{With}}}}
        & \multicolumn{1}{c}{BLEURT} & 0.487 & 0.296 & 0.332 & 0.328 & 0.351\\
        & \multicolumn{1}{c}{COMET(DA)} & 0.495 & 0.290 & 0.315 & 0.336 & 0.328\\
        \cmidrule(l){1-7}
        %& \multicolumn{1}{c}{W/o Human Labels} & Kendall & Pearson & Kendall & Pearson & Kendall & Pearson & Kendall & Pearson\\ \midrule
        \parbox[t]{5mm}{\multirow{7}{*}{\rotatebox[origin=c]{90}{\shortstack{W.o Supervision}}}}
        & \multicolumn{1}{c}{BLEU} & 0.248 & 0.122 & 0.111 & 0.148 & 0.104\\
        & \multicolumn{1}{c}{ChrF} & 0.281 & 0.164 & 0.151 & 0.192 & 0.120\\
        & \multicolumn{1}{c}{TER} & 0.293 & 0.179 & 0.130 & 0.170 & -0.044\\
         & \multicolumn{1}{c}{BERTScore} & 0.411 & 0.264 & 0.223 & 0.247 & 0.269\\
         & \multicolumn{1}{c}{BARTScore} & 0.366 & 0.204 & 0.050 & 0.001 & 0.062\\
         & \multicolumn{1}{c}{PRISM} & 0.396 & 0.257 & 0.192 & 0.310 & 0.230\\
        & \multicolumn{1}{c}{SEScore} & 0.462 & 0.302 & 0.278 & 0.314 & 0.326\\
        & \multicolumn{1}{c}{\method} & \textbf{0.475} & \textbf{0.357} & \textbf{0.298} & \textbf{0.334} & \textbf{0.334}\\
        \bottomrule
    \end{tabular}
    \caption{Segment-level Spearman correlation on En-De and Zh-En for WMT21 News and TED domains. }
    \label{tab:full-lang-results-spearman}
\end{table*}

\begin{table*}[t]\small
    \centering
    \begin{tabular}{@{}l|cccccccc@{}}
        \toprule
        \multicolumn{2}{c}{} & {\bf MT(Zh$\rightarrow$En)} & {\bf MT(En$\rightarrow$De)} & {\bf MT(De$\rightarrow$En)} & {\bf S2T(En$\rightarrow$Ja)} & {\bf D2T} & {\bf Dialogue} & {\bf Overall}\\ \midrule
        \parbox[t]{5mm}{\multirow{2}{*}{\rotatebox[origin=c]{90}{\shortstack{With}}}}
        & \multicolumn{1}{c}{BLEURT} & 0.392 & 0.330 & 0.351 & 0.619 & 0.247 & 0.323 & 0.377 \\
        & \multicolumn{1}{c}{COMET(DA)} & 0.393 & 0.326 & 0.328 & 0.557 & - & - & -\\
        \cmidrule(l){1-9}
        \parbox[t]{5mm}{\multirow{7}{*}{\rotatebox[origin=c]{90}{\shortstack{Without Supervision}}}}
        
        & \multicolumn{1}{c}{TER} & 0.236 & 0.150 & -0.060 & -0.114 & -0.131 & -0.126 & -0.008\\
        & \multicolumn{1}{c}{BLEU} & 0.185 & 0.130 & 0.089 & 0.287 & 0.124 & 0.168 & 0.164\\
        & \multicolumn{1}{c}{ChrF} & 0.223 & 0.172 & 0.098 & 0.336  & 0.139 & 0.168 & 0.189\\
        & \multicolumn{1}{c}{BARTScore} & 0.285 & 0.026 & 0.062 & -0.153 & 0.168 & 0.207 & 0.099\\
        & \multicolumn{1}{c}{BERTScore} & 0.338 & 0.235 & 0.269 & 0.300 & 0.228 & 0.282 & 0.275\\
        & \multicolumn{1}{c}{PRISM} & 0.327 & 0.296 & 0.230 & 0.274 & 0.241 & 0.307 & 0.279\\
        & \multicolumn{1}{c}{SEScore} & 0.382 & 0.296 & 0.326 & 0.493 & 0.228 & 0.298 & 0.337\\
        & \multicolumn{1}{c}{\method} & \textbf{0.416} & \textbf{0.316} & \textbf{0.334} & \textbf{0.616} & \textbf{0.269} & \textbf{0.325} & \textbf{0.379}\\
        \bottomrule
    \end{tabular}
    \caption{Segment-level Spearman correlation on En-De, De-En and Zh-En for WMT21, En-Ja for IWSLT22, WebNLG20 data-to-text and BAGEL dialogue generation.}
    \label{tab:main-text-generation-spearman}
\end{table*}

%% file: paper.bbl
\begin{thebibliography}{46}
\expandafter\ifx\csname natexlab\endcsname\relax\def\natexlab#1{#1}\fi

\bibitem[{Barrault et~al.(2019)Barrault, Bojar, Costa-juss{\`a}, Federmann,
  Fishel, Graham, Haddow, Huck, Koehn, Malmasi, Monz, M{\"u}ller, Pal, Post,
  and Zampieri}]{barrault-etal-2019-findings}
Lo{\"\i}c Barrault, Ond{\v{r}}ej Bojar, Marta~R. Costa-juss{\`a}, Christian
  Federmann, Mark Fishel, Yvette Graham, Barry Haddow, Matthias Huck, Philipp
  Koehn, Shervin Malmasi, Christof Monz, Mathias M{\"u}ller, Santanu Pal, Matt
  Post, and Marcos Zampieri. 2019.
\newblock \href {https://doi.org/10.18653/v1/W19-5301} {Findings of the 2019
  conference on machine translation ({WMT}19)}.
\newblock In \emph{Proceedings of the Fourth Conference on Machine Translation
  (Volume 2: Shared Task Papers, Day 1)}, pages 1--61, Florence, Italy.
  Association for Computational Linguistics.

\bibitem[{Birch(2021)}]{625af6f84b724fcda0a8bf6026cc922f}
Alexandra Birch. 2021.
\newblock \href {https://doi.org/10.1017/S1351324920000650} {Neural machine
  translation 2020, by philipp koehn, cambridge, cambridge university press,
  isbn 978-1-108-49732-9, pages 393.}
\newblock \emph{Natural Language Engineering}, 27(3):377 -- 378.

\bibitem[{Celikyilmaz et~al.(2020)Celikyilmaz, Clark, and
  Gao}]{https://doi.org/10.48550/arxiv.2006.14799}
Asli Celikyilmaz, Elizabeth Clark, and Jianfeng Gao. 2020.
\newblock \href {https://doi.org/10.48550/ARXIV.2006.14799} {Evaluation of text
  generation: A survey}.

\bibitem[{Chopra et~al.(2016)Chopra, Auli, and
  Rush}]{chopra-etal-2016-abstractive}
Sumit Chopra, Michael Auli, and Alexander~M. Rush. 2016.
\newblock \href {https://doi.org/10.18653/v1/N16-1012} {Abstractive sentence
  summarization with attentive recurrent neural networks}.
\newblock In \emph{Proceedings of the 2016 Conference of the North {A}merican
  Chapter of the Association for Computational Linguistics: Human Language
  Technologies}, pages 93--98, San Diego, California. Association for
  Computational Linguistics.

\bibitem[{Chung et~al.(2020)Chung, Fevry, Tsai, Johnson, and
  Ruder}]{chung2020rethinking}
Hyung~Won Chung, Thibault Fevry, Henry Tsai, Melvin Johnson, and Sebastian
  Ruder. 2020.
\newblock Rethinking embedding coupling in pre-trained language models.
\newblock \emph{arXiv preprint arXiv:2010.12821}.

\bibitem[{Conneau et~al.(2019)Conneau, Khandelwal, Goyal, Chaudhary, Wenzek,
  Guzmán, Grave, Ott, Zettlemoyer, and
  Stoyanov}]{https://doi.org/10.48550/arxiv.1911.02116}
Alexis Conneau, Kartikay Khandelwal, Naman Goyal, Vishrav Chaudhary, Guillaume
  Wenzek, Francisco Guzmán, Edouard Grave, Myle Ott, Luke Zettlemoyer, and
  Veselin Stoyanov. 2019.
\newblock \href {https://doi.org/10.48550/ARXIV.1911.02116} {Unsupervised
  cross-lingual representation learning at scale}.

\bibitem[{CONNEAU and Lample(2019)}]{NEURIPS2019_c04c19c2}
Alexis CONNEAU and Guillaume Lample. 2019.
\newblock \href
  {https://proceedings.neurips.cc/paper/2019/file/c04c19c2c2474dbf5f7ac4372c5b9af1-Paper.pdf}
  {Cross-lingual language model pretraining}.
\newblock In \emph{Advances in Neural Information Processing Systems},
  volume~32. Curran Associates, Inc.

\bibitem[{Devlin et~al.(2019)Devlin, Chang, Lee, and
  Toutanova}]{devlin-etal-2019-bert}
Jacob Devlin, Ming-Wei Chang, Kenton Lee, and Kristina Toutanova. 2019.
\newblock \href {https://doi.org/10.18653/v1/N19-1423} {{BERT}: Pre-training of
  deep bidirectional transformers for language understanding}.
\newblock In \emph{Proceedings of the 2019 Conference of the North {A}merican
  Chapter of the Association for Computational Linguistics: Human Language
  Technologies, Volume 1 (Long and Short Papers)}, pages 4171--4186,
  Minneapolis, Minnesota. Association for Computational Linguistics.

\bibitem[{Durmus et~al.(2022)Durmus, Ladhak, and
  Hashimoto}]{durmus-etal-2022-spurious}
Esin Durmus, Faisal Ladhak, and Tatsunori Hashimoto. 2022.
\newblock \href {https://doi.org/10.18653/v1/2022.acl-long.102} {Spurious
  correlations in reference-free evaluation of text generation}.
\newblock In \emph{Proceedings of the 60th Annual Meeting of the Association
  for Computational Linguistics (Volume 1: Long Papers)}, pages 1443--1454,
  Dublin, Ireland. Association for Computational Linguistics.

\bibitem[{Freitag et~al.(2021{\natexlab{a}})Freitag, Foster, Grangier,
  Ratnakar, Tan, and Macherey}]{Freitag_2021}
Markus Freitag, George Foster, David Grangier, Viresh Ratnakar, Qijun Tan, and
  Wolfgang Macherey. 2021{\natexlab{a}}.
\newblock \href {https://doi.org/10.1162/tacl_a_00437} {Experts, errors, and
  context: A large-scale study of human evaluation for machine translation}.
\newblock \emph{Transactions of the Association for Computational Linguistics},
  9:1460--1474.

\bibitem[{Freitag et~al.(2022)Freitag, Grangier, Tan, and
  Liang}]{freitag-etal-2022-high}
Markus Freitag, David Grangier, Qijun Tan, and Bowen Liang. 2022.
\newblock \href {https://doi.org/10.1162/tacl_a_00491} {High quality rather
  than high model probability: Minimum {B}ayes risk decoding with neural
  metrics}.
\newblock \emph{Transactions of the Association for Computational Linguistics},
  10:811--825.

\bibitem[{Freitag et~al.(2021{\natexlab{b}})Freitag, Rei, Mathur, Lo, Stewart,
  Foster, Lavie, and Bojar}]{freitag-etal-2021-results}
Markus Freitag, Ricardo Rei, Nitika Mathur, Chi-kiu Lo, Craig Stewart, George
  Foster, Alon Lavie, and Ond{\v{r}}ej Bojar. 2021{\natexlab{b}}.
\newblock \href {https://aclanthology.org/2021.wmt-1.73} {Results of the
  {WMT}21 metrics shared task: Evaluating metrics with expert-based human
  evaluations on {TED} and news domain}.
\newblock In \emph{Proceedings of the Sixth Conference on Machine Translation},
  pages 733--774, Online. Association for Computational Linguistics.

\bibitem[{Fu et~al.(2022)Fu, Zhou, Xu, Zhou, and Li}]{fu-etal-2022-contextual}
Zhiyi Fu, Wangchunshu Zhou, Jingjing Xu, Hao Zhou, and Lei Li. 2022.
\newblock \href {https://doi.org/10.18653/v1/2022.acl-long.193} {Contextual
  representation learning beyond masked language modeling}.
\newblock In \emph{Proceedings of the 60th Annual Meeting of the Association
  for Computational Linguistics (Volume 1: Long Papers)}, pages 2701--2714,
  Dublin, Ireland. Association for Computational Linguistics.

\bibitem[{Gao et~al.(2021)Gao, Yao, and Chen}]{gao-etal-2021-simcse}
Tianyu Gao, Xingcheng Yao, and Danqi Chen. 2021.
\newblock \href {https://doi.org/10.18653/v1/2021.emnlp-main.552} {{S}im{CSE}:
  Simple contrastive learning of sentence embeddings}.
\newblock In \emph{Proceedings of the 2021 Conference on Empirical Methods in
  Natural Language Processing}, pages 6894--6910, Online and Punta Cana,
  Dominican Republic. Association for Computational Linguistics.

\bibitem[{Gardent et~al.(2017)Gardent, Shimorina, Narayan, and
  Perez-Beltrachini}]{gardent-etal-2017-webnlg}
Claire Gardent, Anastasia Shimorina, Shashi Narayan, and Laura
  Perez-Beltrachini. 2017.
\newblock \href {https://doi.org/10.18653/v1/W17-3518} {The {W}eb{NLG}
  challenge: Generating text from {RDF} data}.
\newblock In \emph{Proceedings of the 10th International Conference on Natural
  Language Generation}, pages 124--133, Santiago de Compostela, Spain.
  Association for Computational Linguistics.

\bibitem[{Graham and Baldwin(2014)}]{graham-baldwin-2014-testing}
Yvette Graham and Timothy Baldwin. 2014.
\newblock \href {https://doi.org/10.3115/v1/D14-1020} {Testing for significance
  of increased correlation with human judgment}.
\newblock In \emph{Proceedings of the 2014 Conference on Empirical Methods in
  Natural Language Processing ({EMNLP})}, pages 172--176, Doha, Qatar.
  Association for Computational Linguistics.

\bibitem[{JTF(2018)}]{JTF}
Japan Translation~Federation JTF. 2018.
\newblock \emph{JTF Translation Quality Evaluation Guidelines}.

\bibitem[{Kepler et~al.(2019)Kepler, Tr{\'e}nous, Treviso, Vera, and
  Martins}]{kepler-etal-2019-openkiwi}
Fabio Kepler, Jonay Tr{\'e}nous, Marcos Treviso, Miguel Vera, and Andr{\'e}
  F.~T. Martins. 2019.
\newblock \href {https://doi.org/10.18653/v1/P19-3020} {{O}pen{K}iwi: An open
  source framework for quality estimation}.
\newblock In \emph{Proceedings of the 57th Annual Meeting of the Association
  for Computational Linguistics: System Demonstrations}, pages 117--122,
  Florence, Italy. Association for Computational Linguistics.

\bibitem[{Koehn(2005)}]{koehn-2005-europarl}
Philipp Koehn. 2005.
\newblock \href {https://aclanthology.org/2005.mtsummit-papers.11} {{E}uroparl:
  A parallel corpus for statistical machine translation}.
\newblock In \emph{Proceedings of Machine Translation Summit X: Papers}, pages
  79--86, Phuket, Thailand.

\bibitem[{Kryscinski et~al.(2020)Kryscinski, McCann, Xiong, and
  Socher}]{kryscinski-etal-2020-evaluating}
Wojciech Kryscinski, Bryan McCann, Caiming Xiong, and Richard Socher. 2020.
\newblock \href {https://doi.org/10.18653/v1/2020.emnlp-main.750} {Evaluating
  the factual consistency of abstractive text summarization}.
\newblock In \emph{Proceedings of the 2020 Conference on Empirical Methods in
  Natural Language Processing (EMNLP)}, pages 9332--9346, Online. Association
  for Computational Linguistics.

\bibitem[{K{\'{u}}dela et~al.(2017)K{\'{u}}dela, Holubov{\'{a}}, and
  Bojar}]{K_dela_2017}
Jakub K{\'{u}}dela, Irena Holubov{\'{a}}, and Ond{\v{r}}ej Bojar. 2017.
\newblock \href {https://doi.org/10.1515/pralin-2017-0003} {Extracting parallel
  paragraphs from common crawl}.
\newblock \emph{The Prague Bulletin of Mathematical Linguistics},
  107(1):39--56.

\bibitem[{Liu et~al.(2020)Liu, Gu, Goyal, Li, Edunov, Ghazvininejad, Lewis, and
  Zettlemoyer}]{liu-etal-2020-multilingual-denoising}
Yinhan Liu, Jiatao Gu, Naman Goyal, Xian Li, Sergey Edunov, Marjan
  Ghazvininejad, Mike Lewis, and Luke Zettlemoyer. 2020.
\newblock \href {https://doi.org/10.1162/tacl_a_00343} {Multilingual denoising
  pre-training for neural machine translation}.
\newblock \emph{Transactions of the Association for Computational Linguistics},
  8:726--742.

\bibitem[{Liu et~al.(2019)Liu, Ott, Goyal, Du, Joshi, Chen, Levy, Lewis,
  Zettlemoyer, and Stoyanov}]{Liu2019RoBERTaAR}
Yinhan Liu, Myle Ott, Naman Goyal, Jingfei Du, Mandar Joshi, Danqi Chen, Omer
  Levy, Mike Lewis, Luke Zettlemoyer, and Veselin Stoyanov. 2019.
\newblock Roberta: A robustly optimized bert pretraining approach.
\newblock \emph{ArXiv}, abs/1907.11692.

\bibitem[{Lommel et~al.(2014)Lommel, Uszkoreit, and Burchardt}]{lommel2014mqm}
Arle Lommel, Hans Uszkoreit, and Aljoscha Burchardt. 2014.
\newblock Mqm: Un marc per declarar i descriure m{\`e}triques de qualitat de la
  traducci{\'o}.
\newblock \emph{Tradum{\`a}tica: traducci{\'o} i tecnologies de la
  informaci{\'o} i la comunicaci{\'o}}, (12):455--463.

\bibitem[{Louis and Nenkova(2013)}]{louis-nenkova-2013-automatically}
Annie Louis and Ani Nenkova. 2013.
\newblock \href {https://doi.org/10.1162/COLI_a_00123} {Automatically assessing
  machine summary content without a gold standard}.
\newblock \emph{Computational Linguistics}, 39(2):267--300.

\bibitem[{Ma et~al.(2018)Ma, Bojar, and Graham}]{ma-etal-2018-results}
Qingsong Ma, Ond{\v{r}}ej Bojar, and Yvette Graham. 2018.
\newblock \href {https://doi.org/10.18653/v1/W18-6450} {Results of the {WMT}18
  metrics shared task: Both characters and embeddings achieve good
  performance}.
\newblock In \emph{Proceedings of the Third Conference on Machine Translation:
  Shared Task Papers}, pages 671--688, Belgium, Brussels. Association for
  Computational Linguistics.

\bibitem[{Ma et~al.(2019)Ma, Wei, Bojar, and Graham}]{ma-etal-2019-results}
Qingsong Ma, Johnny Wei, Ond{\v{r}}ej Bojar, and Yvette Graham. 2019.
\newblock \href {https://doi.org/10.18653/v1/W19-5302} {Results of the {WMT}19
  metrics shared task: Segment-level and strong {MT} systems pose big
  challenges}.
\newblock In \emph{Proceedings of the Fourth Conference on Machine Translation
  (Volume 2: Shared Task Papers, Day 1)}, pages 62--90, Florence, Italy.
  Association for Computational Linguistics.

\bibitem[{Mathur et~al.(2020)Mathur, Wei, Freitag, Ma, and
  Bojar}]{mathur-etal-2020-results}
Nitika Mathur, Johnny Wei, Markus Freitag, Qingsong Ma, and Ond{\v{r}}ej Bojar.
  2020.
\newblock \href {https://aclanthology.org/2020.wmt-1.77} {Results of the
  {WMT}20 metrics shared task}.
\newblock In \emph{Proceedings of the Fifth Conference on Machine Translation},
  pages 688--725, Online. Association for Computational Linguistics.

\bibitem[{Papineni et~al.(2002)Papineni, Roukos, Ward, and
  Zhu}]{papineni-etal-2002-bleu}
Kishore Papineni, Salim Roukos, Todd Ward, and Wei-Jing Zhu. 2002.
\newblock \href {https://doi.org/10.3115/1073083.1073135} {{B}leu: a method for
  automatic evaluation of machine translation}.
\newblock In \emph{Proceedings of the 40th Annual Meeting of the Association
  for Computational Linguistics}, pages 311--318, Philadelphia, Pennsylvania,
  USA. Association for Computational Linguistics.

\bibitem[{Popovi{\'c}(2015)}]{popovic-2015-chrf}
Maja Popovi{\'c}. 2015.
\newblock \href {https://doi.org/10.18653/v1/W15-3049} {chr{F}: character
  n-gram {F}-score for automatic {MT} evaluation}.
\newblock In \emph{Proceedings of the Tenth Workshop on Statistical Machine
  Translation}, pages 392--395, Lisbon, Portugal. Association for Computational
  Linguistics.

\bibitem[{Pryzant et~al.(2018)Pryzant, Chung, Jurafsky, and
  Britz}]{pryzant-etal-2018-jesc}
Reid Pryzant, Youngjoo Chung, Dan Jurafsky, and Denny Britz. 2018.
\newblock \href {https://aclanthology.org/L18-1182} {{JESC}:
  {J}apanese-{E}nglish subtitle corpus}.
\newblock In \emph{Proceedings of the Eleventh International Conference on
  Language Resources and Evaluation ({LREC} 2018)}, Miyazaki, Japan. European
  Language Resources Association (ELRA).

\bibitem[{Rei et~al.(2020)Rei, Stewart, Farinha, and
  Lavie}]{rei-etal-2020-comet}
Ricardo Rei, Craig Stewart, Ana~C Farinha, and Alon Lavie. 2020.
\newblock \href {https://doi.org/10.18653/v1/2020.emnlp-main.213} {{COMET}: A
  neural framework for {MT} evaluation}.
\newblock In \emph{Proceedings of the 2020 Conference on Empirical Methods in
  Natural Language Processing (EMNLP)}, pages 2685--2702, Online. Association
  for Computational Linguistics.

\bibitem[{Schwenk et~al.(2021)Schwenk, Chaudhary, Sun, Gong, and
  Guzm{\'a}n}]{schwenk-etal-2021-wikimatrix}
Holger Schwenk, Vishrav Chaudhary, Shuo Sun, Hongyu Gong, and Francisco
  Guzm{\'a}n. 2021.
\newblock \href {https://doi.org/10.18653/v1/2021.eacl-main.115}
  {{W}iki{M}atrix: Mining 135{M} parallel sentences in 1620 language pairs from
  {W}ikipedia}.
\newblock In \emph{Proceedings of the 16th Conference of the European Chapter
  of the Association for Computational Linguistics: Main Volume}, pages
  1351--1361, Online. Association for Computational Linguistics.

\bibitem[{Sellam et~al.(2020)Sellam, Das, and Parikh}]{sellam-etal-2020-bleurt}
Thibault Sellam, Dipanjan Das, and Ankur Parikh. 2020.
\newblock \href {https://doi.org/10.18653/v1/2020.acl-main.704} {{BLEURT}:
  Learning robust metrics for text generation}.
\newblock In \emph{Proceedings of the 58th Annual Meeting of the Association
  for Computational Linguistics}, pages 7881--7892, Online. Association for
  Computational Linguistics.

\bibitem[{Shimanaka et~al.(2018)Shimanaka, Kajiwara, and
  Komachi}]{shimanaka-etal-2018-ruse}
Hiroki Shimanaka, Tomoyuki Kajiwara, and Mamoru Komachi. 2018.
\newblock \href {https://doi.org/10.18653/v1/W18-6456} {{RUSE}: Regressor using
  sentence embeddings for automatic machine translation evaluation}.
\newblock In \emph{Proceedings of the Third Conference on Machine Translation:
  Shared Task Papers}, pages 751--758, Belgium, Brussels. Association for
  Computational Linguistics.

\bibitem[{Snover et~al.(2006)Snover, Dorr, Schwartz, Micciulla, and
  Makhoul}]{snover-etal-2006-study}
Matthew Snover, Bonnie Dorr, Rich Schwartz, Linnea Micciulla, and John Makhoul.
  2006.
\newblock \href {https://aclanthology.org/2006.amta-papers.25} {A study of
  translation edit rate with targeted human annotation}.
\newblock In \emph{Proceedings of the 7th Conference of the Association for
  Machine Translation in the Americas: Technical Papers}, pages 223--231,
  Cambridge, Massachusetts, USA. Association for Machine Translation in the
  Americas.

\bibitem[{Thompson and Post(2020)}]{thompson-post-2020-automatic}
Brian Thompson and Matt Post. 2020.
\newblock \href {https://doi.org/10.18653/v1/2020.emnlp-main.8} {Automatic
  machine translation evaluation in many languages via zero-shot paraphrasing}.
\newblock In \emph{Proceedings of the 2020 Conference on Empirical Methods in
  Natural Language Processing (EMNLP)}, pages 90--121, Online. Association for
  Computational Linguistics.

\bibitem[{Tiedemann(2012)}]{tiedemann-2012-parallel}
J{\"o}rg Tiedemann. 2012.
\newblock \href
  {http://www.lrec-conf.org/proceedings/lrec2012/pdf/463_Paper.pdf} {Parallel
  data, tools and interfaces in {OPUS}}.
\newblock In \emph{Proceedings of the Eighth International Conference on
  Language Resources and Evaluation ({LREC}'12)}, pages 2214--2218, Istanbul,
  Turkey. European Language Resources Association (ELRA).

\bibitem[{Unanue et~al.(2021)Unanue, Parnell, and
  Piccardi}]{JauregiUnanue2021BERTTuneFN}
Inigo~Jauregi Unanue, Jacob Parnell, and Massimo Piccardi. 2021.
\newblock Berttune: Fine-tuning neural machine translation with bertscore.
\newblock In \emph{ACL}.

\bibitem[{Vinyals and Le(2015)}]{https://doi.org/10.48550/arxiv.1506.05869}
Oriol Vinyals and Quoc Le. 2015.
\newblock \href {https://doi.org/10.48550/ARXIV.1506.05869} {A neural
  conversational model}.

\bibitem[{Xu et~al.(2022)Xu, Tuan, Lu, Saxon, Li, and Wang}]{xu-etal-2022-not}
Wenda Xu, Yi-lin Tuan, Yujie Lu, Michael Saxon, Lei Li, and William~Yang Wang.
  2022.
\newblock \href {https://arxiv.org/abs/2210.05035} {Not all errors are equal:
  Learning text generation metrics using stratified error synthesis}.
\newblock In \emph{Proceedings of the 2022 Conference on Empirical Methods in
  Natural Language Processing}.

\bibitem[{Yuan et~al.(2021)Yuan, Neubig, and
  Liu}]{https://doi.org/10.48550/arxiv.2106.11520}
Weizhe Yuan, Graham Neubig, and Pengfei Liu. 2021.
\newblock \href {https://doi.org/10.48550/ARXIV.2106.11520} {Bartscore:
  Evaluating generated text as text generation}.

\bibitem[{Zhang et~al.(2019)Zhang, Kishore, Wu, Weinberger, and
  Artzi}]{https://doi.org/10.48550/arxiv.1904.09675}
Tianyi Zhang, Varsha Kishore, Felix Wu, Kilian~Q. Weinberger, and Yoav Artzi.
  2019.
\newblock \href {https://doi.org/10.48550/ARXIV.1904.09675} {Bertscore:
  Evaluating text generation with bert}.

\bibitem[{Zhao et~al.(2019)Zhao, Peyrard, Liu, Gao, Meyer, and
  Eger}]{zhao-etal-2019-moverscore}
Wei Zhao, Maxime Peyrard, Fei Liu, Yang Gao, Christian~M. Meyer, and Steffen
  Eger. 2019.
\newblock \href {https://doi.org/10.18653/v1/D19-1053} {{M}over{S}core: Text
  generation evaluating with contextualized embeddings and earth mover
  distance}.
\newblock In \emph{Proceedings of the 2019 Conference on Empirical Methods in
  Natural Language Processing and the 9th International Joint Conference on
  Natural Language Processing (EMNLP-IJCNLP)}, pages 563--578, Hong Kong,
  China. Association for Computational Linguistics.

\bibitem[{Zhou and Lampouras(2020)}]{zhou-lampouras-2020-webnlg}
Giulio Zhou and Gerasimos Lampouras. 2020.
\newblock \href {https://aclanthology.org/2020.webnlg-1.22} {{W}eb{NLG}
  challenge 2020: Language agnostic delexicalisation for multilingual
  {RDF}-to-text generation}.
\newblock In \emph{Proceedings of the 3rd International Workshop on Natural
  Language Generation from the Semantic Web (WebNLG+)}, pages 186--191, Dublin,
  Ireland (Virtual). Association for Computational Linguistics.

\bibitem[{Ziemski et~al.(2016)Ziemski, Junczys-Dowmunt, and
  Pouliquen}]{ziemski-etal-2016-united}
Micha{\l} Ziemski, Marcin Junczys-Dowmunt, and Bruno Pouliquen. 2016.
\newblock \href {https://aclanthology.org/L16-1561} {The {U}nited {N}ations
  parallel corpus v1.0}.
\newblock In \emph{Proceedings of the Tenth International Conference on
  Language Resources and Evaluation ({LREC}'16)}, pages 3530--3534,
  Portoro{\v{z}}, Slovenia. European Language Resources Association (ELRA).

\end{thebibliography}
